\documentclass{article}
\PassOptionsToPackage{numbers,sort&compress}{natbib}
\usepackage[preprint]{neurips_2026}
\usepackage[T1]{fontenc}
\usepackage[utf8]{inputenc}
\usepackage{graphicx}
\usepackage{multirow}
\usepackage{amsmath,amssymb,amsfonts,amsthm,mathrsfs}
\usepackage[title]{appendix}
\usepackage{xcolor,textcomp,manyfoot,booktabs}
\usepackage{algorithm,algorithmicx,algpseudocode,listings}
\usepackage{url}
\usepackage[hidelinks]{hyperref}
\usepackage{adjustbox}
\usepackage{float}
\usepackage[section]{placeins}

\let\originalincludegraphics\includegraphics
\renewcommand{\includegraphics}[2][]{%
  \adjustbox{max totalheight=.70\textheight}{\originalincludegraphics[#1]{#2}}}
\let\originalAA\AA
\renewcommand{\AA}{\ifmmode\mathring{\mathrm{A}}\else\originalAA\fi}

\theoremstyle{plain}

\theoremstyle{remark}

\theoremstyle{definition}

\newcommand{\backmatter}{%
  \let\savedfigure\figure
  \renewcommand{\figure}[1][tbp]{\savedfigure[!htbp]}}

\title{Structural Agreement Indicates Prediction Accuracy For Protein-Ligand Co-Folding}
\author{\parbox{\dimexpr\textwidth-2\tabcolsep\relax}{\centering
J. Henry Broster\textsuperscript{1}\quad
Bojana Popovic\textsuperscript{2}\quad
Diana Kondinskaia\textsuperscript{2}\\[4pt]
Charlotte M. Deane\textsuperscript{1}\quad
Fergus Imrie\textsuperscript{1}\\[8pt]
\normalfont\small
\textsuperscript{1}Department of Statistics, University of Oxford,\\
Oxford, OX1 3LB, United Kingdom\\[3pt]
\textsuperscript{2}Cambridge Crystallographic Data Centre,\\
Cambridge, CB2 1EZ, United Kingdom\\[6pt]
\texttt{broster@stats.ox.ac.uk}\quad\texttt{bpopovic@ccdc.cam.ac.uk}\\
\texttt{dkondinskaia@ccdc.cam.ac.uk}\quad\texttt{deane@stats.ox.ac.uk}\\
\texttt{imrie@stats.ox.ac.uk}
}}
\begin{document}
\maketitle
\begin{abstract}
Co-folding models predict the structures of proteins and their bound ligands, but their accuracy varies across targets. Their usefulness therefore depends on knowing which predictions a user can trust. Confidence scores are intended to guide that choice. However, across seven co-folding models, we show that confidence-based selection offers little improvement over choosing a prediction at random. For approximately three-quarters of structures, predictions are either exclusively correct or incorrect, leaving no opportunity for ranking to change the outcome. Moreover, high confidence does not reliably discriminate correct top-ranked predictions from incorrect ones. We find that structural agreement among predictions is a more accurate measure of likely success than current confidence scores. We capture this with the Concentration of Repeated Ensembles (CORE) Score, which combines cluster tightness with ensemble support. CORE Score discriminates top-1 success and oracle success, achieving mean ROC AUCs of 0.866 and 0.838, compared to 0.757 and 0.759 for ipTM. CORE Score also outperforms training set similarity measures, including ligand Tanimoto, protein sequence, and Runs N' Poses similarity, despite requiring no experimental ligand pose. Structural agreement is therefore a practical indicator of reliability and a baseline for evaluating future confidence estimators, without requiring an additional trained uncertainty estimator.
\end{abstract}

\section{Introduction}

Predicting the three-dimensional structure of a protein–ligand complex is central to structure-based drug design \cite{batool_structure-based_2019}. The computational study of protein-ligand binding has historically been addressed by molecular docking \cite{meng_molecular_2011, agu_molecular_2023}, which has often treated the receptor as rigid, predicting possible positions and conformations of a ligand within a plausible binding site. A scoring function is then used both to rank the generated poses for a given complex and to assess whether the ligand is likely to bind the target \cite{trott_autodock_2010, friesner_glide_2004}. Docked poses underpin virtual screening and lead optimisation \cite{pinzi_molecular_2019}, but prospective performance varies substantially across targets and the problem remains unsolved \cite{buttenschoen_posebusters_2024, skrinjar_have_2025}.

Deep learning has enabled diffusion-based docking models that generate ligand poses conditioned on a supplied protein structure \cite{corso2024deep, broster_teaching_2026}. More recently, a class of diffusion-based models has been developed, termed co-folding models, which predict the protein and the ligand jointly using the protein sequence and ligand structure as input \cite{abramson_accurate_2024, wohlwend_boltz-1_2024, chai_discovery_chai-1_2024, bytedance_aml_ai4science_team_protenix_2025, openfold3-preview}. In doing so, they bring together protein structure prediction and ligand pose prediction, two problems that had previously been treated as separate, into a single generative task. 

Despite advances in co-folding, determining which predictions can be trusted remains a central challenge for prospective use \cite{skrinjar_have_2025, xu_benchmarking_2025}. Benchmark success rates describe performance on average, but they do not indicate whether a particular prediction is accurate. The practical value of these models therefore depends on identifying when their structural predictions are likely to be correct.

While prediction accuracy has been shown to depend strongly on how similar a complex is to the training data \cite{skrinjar_have_2025}, protein sequence identity and ligand Morgan fingerprint Tanimoto similarity \cite{rogers_extended-connectivity_2010} do not consider the binding mode itself. Runs N' Poses therefore defined similarity using both the ligand pose and its binding pocket, so that a training structure is considered similar if it has a similar ligand pose in a similar protein pocket \cite{skrinjar_have_2025}. However, because this metric requires the experimental ligand pose, it cannot be calculated when the true structure is unknown. Prospective assessment therefore requires indicators that reliably distinguish accurate from inaccurate predictions using only the supplied inputs, generation process, or predictions themselves. 

Following earlier structure prediction and docking models such as AlphaFold2 and DiffDock \cite{jumper_highly_2021, corso2023diffdock, broster_teaching_2026}, co-folding models provide such an assessment through learned confidence estimates \cite{abramson_accurate_2024}. For instance, predicted local distance difference test (pLDDT) scores estimate local structural accuracy \cite{jumper_highly_2021}, while interface predicted TM-scores (ipTM) assess interface accuracy, including protein–ligand interfaces \cite{abramson_accurate_2024}. These confidence scores are used to inform which predictions are more or less likely to be structurally accurate, and therefore serve two distinct purposes: (1) to rank predictions for a given complex and identify the pose most likely to be correct, and (2) to distinguish predictions which are correct from those which are incorrect. Beyond structural correctness, confidence scores have been assessed in a virtual screening setting \cite{Kim_2026}. Here, we focus on the accuracy of the predicted binding pose, rather than distinguishing between binders and non-binders.

\begin{figure}[t!]
    \centering
    \includegraphics[width=\textwidth]{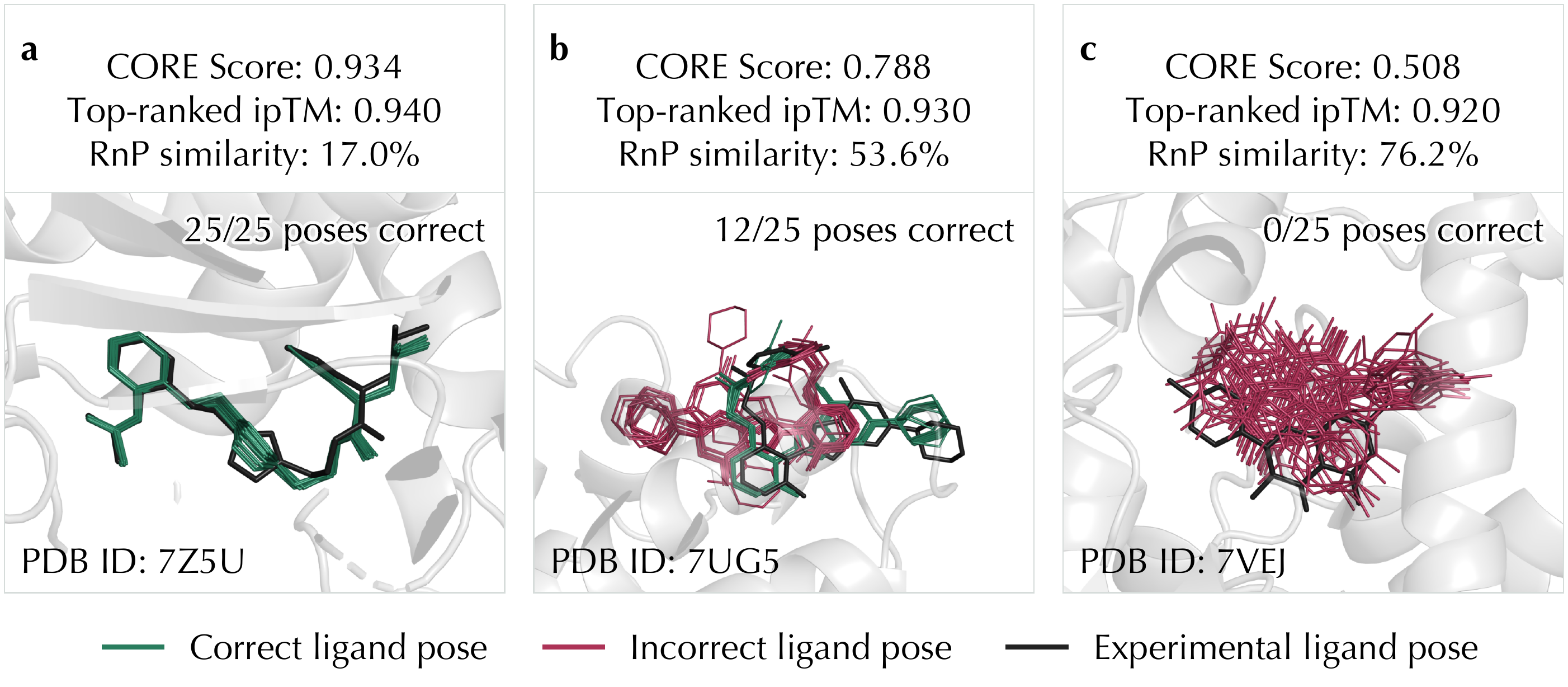}
    \caption{
    \textbf{Structural agreement reflects prediction accuracy in illustrative ensembles.}
    Three AlphaFold3 ensembles, each containing 25 predicted structures, illustrate (a) entirely correct, (b) mixed, and (c) entirely incorrect prediction ensembles. Across these examples, CORE Score falls as the proportion of correct poses falls, whereas similarity to training examples rises and top-ranked mean chain-pair ipTM remains high for all three. Green and red indicate correct and incorrect ligand poses, respectively. The experimental ligand is shown in black and the reference protein in grey.
    }
    \label{fig:introduction}
\end{figure}

Diffusion-based co-folding models generate structures by sampling from a learned conditional distribution \cite{ho_denoising_2020,abramson_accurate_2024}. Repeated predictions can be used to characterise this distribution empirically, revealing how the model distributes its support across possible structural solutions for a given complex. A less diverse set of predictions suggests high predictive certainty, with the model concentrating its support around a single structural solution. A more diverse prediction ensemble suggests greater predictive uncertainty, with support spread across a range of structures or competing solutions (Figure~\ref{fig:introduction}). Unlike learned confidence estimates, this requires no training and uses only the generated structures themselves. We hypothesised that this concentration would be associated with prediction correctness, with more concentrated ensembles more likely to contain correct poses. However, since a model may concentrate around an incorrect solution, certainty does not guarantee correctness. We therefore investigate whether structural agreement indicates prediction correctness across seven co-folding models, and whether this provides information beyond learned confidence scores and training set similarity \cite{skrinjar_have_2025}.

We begin by investigating learned confidence scores. We find that selecting the highest-ipTM pose within ensembles where there are both correct and incorrect predictions offers only modest improvement over random selection. When assessing whether a top-ranked prediction is correct, ipTM provides some separation between correct and incorrect predictions, but their distributions overlap substantially, limiting its reliability for identifying successful predictions. Examining prediction diversity, we observe a pronounced separation between outcomes, with highly concentrated ensembles consistent with mode collapse overwhelmingly containing only correct poses, while ensembles with low concentration overwhelmingly contain only incorrect poses. The same relationship appears within ensembles, where larger and tighter clusters contain a higher proportion of correct poses. These observations inform the Concentration of Repeated Ensembles (CORE) Score, which combines cluster tightness with ensemble support. To reflect a prospective, real-world scenario, we remove physically invalid poses and find that CORE Score discriminates top-1 success with a mean ROC AUC of 0.866, compared with 0.757 for ipTM, and oracle success with a mean ROC AUC of 0.838, compared with 0.759 for ipTM. CORE Score also outperforms protein sequence and ligand Tanimoto similarity, as well as the Runs N’ Poses similarity metric across all seven models, despite CORE Score requiring no experimental ligand pose. These results support ensemble concentration as a practical indicator of prediction quality for prospective applications and a baseline against which learned confidence estimators can be evaluated.

\section{Methods}

\subsection{Runs N' Poses Benchmark and Co-Folding Predictions}

We evaluated the relationship between structural agreement and prediction accuracy using predicted structures from the Runs N’ Poses benchmark \cite{skrinjar_have_2025}, which comprises 2,600 high-resolution protein–ligand systems derived from the Protein Data Bank (PDB) \cite{berman_protein_2000}. Our analysis includes predictions from AlphaFold3 \cite{abramson_accurate_2024}, Boltz-1 \cite{wohlwend_boltz-1_2024}, Boltz-1x, Boltz-2 \cite{passaro_boltz-2_2025}, Chai-1 \cite{chai_discovery_chai-1_2024}, Protenix \cite{bytedance_aml_ai4science_team_protenix_2025}, and OpenFold3 p2 \cite{openfold3-preview}. We analysed these structures together with their model-specific overall confidence scores and the ligand-protein chain-pair ipTM value for each predicted pose, as well as reference ligand RMSD and lDDT-PLI measurements, PoseBusters-validity checks, and three measures of similarity to the training set: protein sequence identity, ligand Morgan fingerprint Tanimoto similarity \cite{rogers_extended-connectivity_2010}, and the Runs N’ Poses similarity metric \cite{skrinjar_have_2025}. The overall confidence score and ipTM differ in that ipTM assesses the predicted accuracy of the interface between the ligand and a protein chain, whereas the overall confidence score is a single model-specific value summarising each predicted structure.

Analyses were restricted to single copies of the ligand being evaluated, excluding complexes containing multiple copies of that ligand (Figures~\ref{fig:si_7wl4} and~\ref{fig:si_multi_examples}). We retained all ensembles with at least one prediction record having \texttt{ligand\_is\_proper=True} for the corresponding pose in the annotation files. Retention varied between models because the released predictions cover different subsets of the benchmark, so the proportion of targets with multiple copies of the evaluated ligand differed accordingly. These restrictions retained 79.2\% to 85.0\% of available ensembles across models, with a mean of 81.8\%. The retained dataset comprises 13,396 model-specific ensembles representing 2,371 distinct target-ligand inputs, each containing 25 generated poses. Per-model counts are provided in Table~\ref{tab:si_coverage}. For comparisons involving training-set similarity, we additionally excluded 406 ensembles missing protein sequence identity, ligand Tanimoto similarity, or Runs N' Poses similarity annotations. Together, these restrictions retained 74.8\% to 83.1\% of available ensembles across models, with a mean of 79.5\%. The resulting comparison dataset comprises 12,990 model-specific ensembles representing 2,212 distinct target-ligand inputs.

\subsection{Evaluation of Prediction Reliability}

Following Runs N' Poses \cite{skrinjar_have_2025}, a prediction was considered correct if the RMSD of the predicted ligand to the experimental pose was below $2$~\AA\ and its lDDT-PLI, measuring how well the protein-ligand contacts of the experimental structure are reproduced \cite{robin_assessment_2023}, was above $0.8$. Initial descriptive analyses retained all poses. For prospective reliability analyses, we removed PoseBusters-invalid poses before selecting the top-ranked prediction, so that success additionally required PoseBusters validity \cite{buttenschoen_posebusters_2024}.

We compared CORE Score with mean ligand-protein chain-pair ipTM, following Runs N' Poses \cite{skrinjar_have_2025}, as well as protein sequence identity, ligand Tanimoto similarity, and the Runs N' Poses similarity metric to training examples. Of these, only the Runs N' Poses similarity metric requires the experimental ligand pose, so it cannot be calculated prospectively and is included as a reference. For CORE Score, ipTM, and similarity comparisons, the top-ranked prediction was the pose with the highest mean ligand-protein chain-pair ipTM. For overall confidence comparisons, the top-ranked pose was the one with the highest model-specific overall confidence score. Discrimination was assessed using ROC AUC for top-1 success (whether the top-ranked prediction was correct) and oracle success (whether any prediction in the ensemble was correct). Statistical significance of differences between measures, and 95\% confidence intervals on the equal-model mean ROC AUC, were assessed using a paired cluster bootstrap over PDB entries with 2,000 replicates \cite{field_bootstrapping_2007}.

\subsection{Quantifying Structural Agreement}
\label{sec:quantifying}

Repeated predictions for the same protein-ligand input sample the model’s learned conditional distribution. Their spread characterises where the model places its support, through both the number of predictions recovering a given structure and the similarity between them. We hypothesised that repeated recovery of closely related poses would indicate a greater likelihood of a correct prediction, without requiring comparison with training examples. 

We primarily capture structural agreement with pairwise ligand RMSDs. Pairwise ligand RMSDs were calculated after binding-site superposition with OpenStructure \cite{ost_2026}, implementation details of which are given in Section~\ref{sec:si_methods}.

An ensemble of 25 poses gives 300 unique pairwise RMSD values. As simple baselines, we calculate the minimum pairwise RMSD, the mean pairwise RMSD and the median pairwise RMSD. Lower values indicate closer agreement between poses.

\textbf{CORE Score.} The observation that larger, tighter groups generally contained higher proportions of correct predictions motivated the Concentration of Repeated Ensembles (CORE) Score. CORE Score measures how closely each generated pose is reproduced by any other pose in the ensemble. For each pose $i$ in an ensemble of $N$ poses, we took the ligand RMSD to each of its three closest other poses,
\begin{equation}
    \delta_{i,m} = \min^{(m)}_{j \neq i} d_{ij}, \quad m = 1, 2, 3,
    \label{eq:concentration_distance}
\end{equation}
where $d_{ij}$ is the pairwise ligand RMSD between poses $i$ and $j$ and $\min^{(m)}$ denotes the $m$th smallest value. Each distance was capped at a threshold $T$ and divided by $T$, giving a penalty of zero for a pose reproduced exactly by its three closest poses and one for a pose with no other pose within $T$. CORE Score $S$ is given as one minus the mean penalty across the ensemble:
\begin{equation}
    S = 1 - \frac{1}{3N}\sum_{i=1}^{N}\sum_{m=1}^{3} \frac{\min(\delta_{i,m}, T)}{T}.
    \label{eq:concentration_score}
\end{equation}
We set $T$ equal to the RMSD threshold used to define a correct prediction, $T = 2$~\AA. Results for stricter success thresholds, with $T$ set to match, are given in Tables~\ref{tab:si_threshold} and~\ref{tab:si_threshold_grid}. Distances beyond $T$ were treated equally, since we view a pose with no other pose within $T$ as equally unsupported by the ensemble. Dividing by $T$ places CORE Score between zero and one without changing the ordering of ensembles.

We can equivalently write $S = f\,(1 - \nu/T)$, where $f$ is the fraction of closest-pose distances $\delta_{i,m}$ within $T$ and $\nu$ is the mean of those distances. This follows by separating the capped distances into those within $T$, which contribute $\nu/T$ on average, and those at or beyond $T$, which each contribute one. CORE Score therefore increases with both the fraction of nearest-neighbour distances within $T$ and how small those distances are, capturing ensemble support and tightness without requiring clustering. Using the three closest poses means that a pose scores highly only when at least three other poses lie close to it, so a high CORE Score requires groups of at least four closely agreeing poses rather than isolated matching pairs. The ensemble as a whole scores highly only when most of its poses belong to such groups.

CORE Score was calculated from all 25 generated poses, including those that failed PoseBusters checks, as these invalid poses still sample the model’s output distribution. Validity filtering was applied only when selecting the top-ranked prediction, which was the PoseBusters-valid pose with the highest ligand-protein chain-pair ipTM. Ensembles containing no PoseBusters-valid pose provide no prediction to evaluate and were excluded rather than scored, accounting for 6.2\% of ensembles. Since PoseBusters validity requires no experimental structure, these ensembles are identified before any prediction is selected, as they would be in prospective use. The baseline comparisons followed the same protocol (Figures~\ref{fig:si_valid_alternatives_auc} and~\ref{fig:si_allposes_auc}).

\section{Results}

\FloatBarrier
\subsection{Confidence Scores Incompletely Capture Prediction Correctness}

\begin{figure}[!t]
    \centering
    \includegraphics[width=0.9\textwidth]{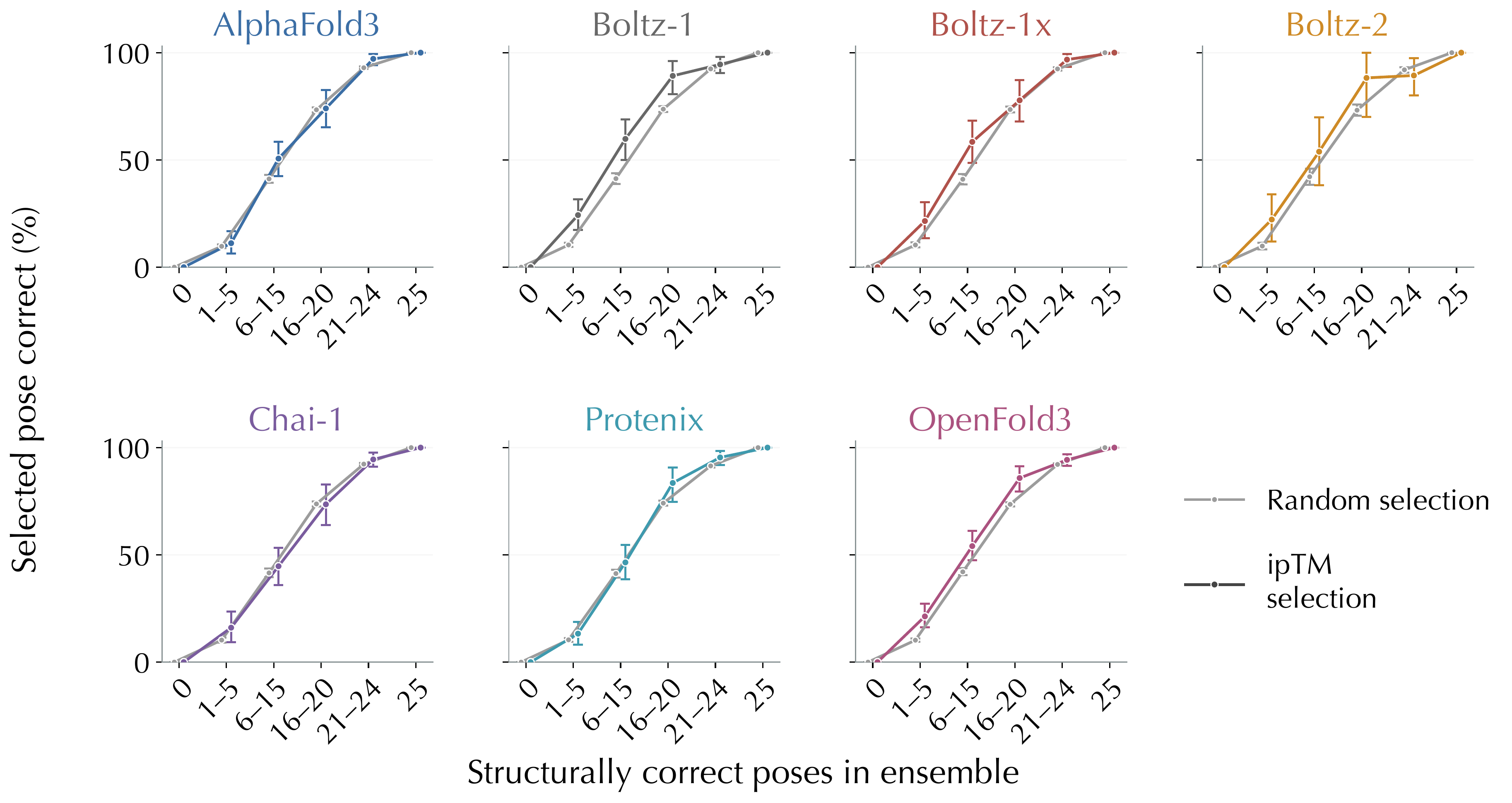}
    \caption{\textbf{Existing confidence metrics provide limited performance for pose selection.} Highest-ipTM selection is compared with uniform random selection across ensembles grouped by the number of correct poses in an ensemble. Points show observed success rates and bars show 95\% bootstrap intervals from resampling PDB entries.}
    \label{fig:confidence}
\end{figure}

We evaluated ligand-protein chain-pair ipTM in two settings: (1) whether confidence could rank correct poses above incorrect alternatives for the same complex, and (2) whether confidence could distinguish correct top-ranked predictions from incorrect ones.

To assess whether confidence scores could identify a correct pose among the generated predictions, we stratified by the number of correct poses in an ensemble and compared selection of the top-ranked pose as judged using ligand-protein chain-pair ipTM with random selection across ensembles. This allows us to isolate how well the confidence model performs relative to the number of correct poses in a batch. Within these ensembles, the fraction of correct poses defines the expected success rate of random selection.

We observed that, across all models, confidence-based selection offered limited to no improvement over the random baseline. Averaged equally across models, the mean success rate was 60.4\% for ipTM-based selection compared with 58.6\% for random selection, an improvement of only 1.8\%. This shows that ipTM fails to reliably prioritise correct poses over incorrect alternatives (Figures~\ref{fig:confidence} and~\ref{fig:si_sel_native}). 

\begin{figure}[!t]
    \centering
    \includegraphics[width=0.9\textwidth]{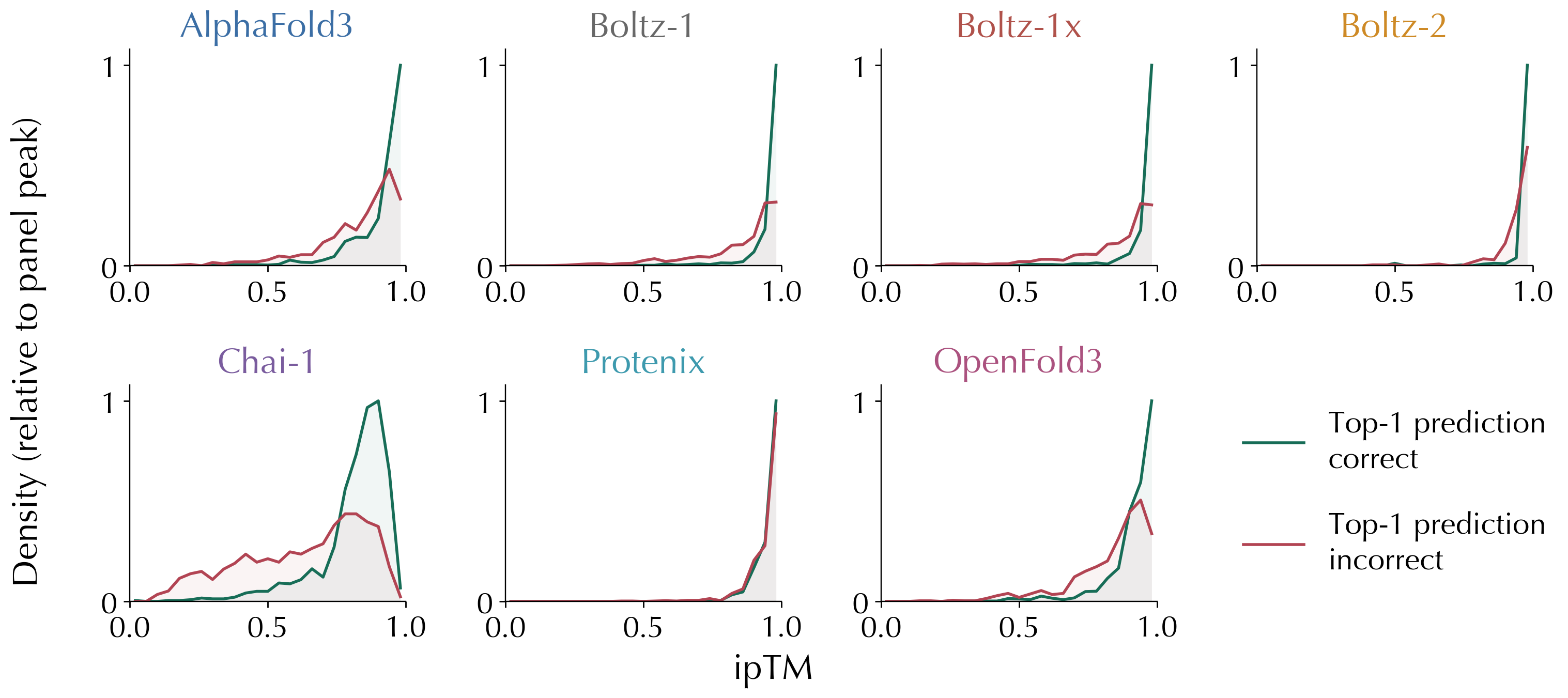}
    \caption{\textbf{Correct and incorrect top predictions have overlapping confidence distributions.} The highest-ipTM prediction is selected for each model, and its ipTM value is shown. Green and red distributions are normalised separately within the correct and incorrect outcomes. Correctness uses RMSD and lDDT-PLI without PB filtering.}
    \label{fig:confidence_distributions}
\end{figure}

However, the limited benefit of confidence-based ranking does not necessarily mean that confidence is uninformative about correctness. A score can be unable to rank poses within an ensemble while still distinguishing correct top-ranked predictions from incorrect ones. To assess this, we compared the ipTM distributions of correct and incorrect highest-ipTM predictions separately for each model (Figures~\ref{fig:confidence_distributions},~\ref{fig:si_dist_native} and~\ref{fig:si_core_dist}). While the degree of separation varies between models, these distributions overlap substantially. Quantified by ROC AUC, ipTM discrimination of top-1 success ranges from 0.612 to 0.839 across models, with a mean of 0.757. High confidence alone therefore does not reliably distinguish correct predictions from incorrect ones.

\subsection{Ensembles Frequently Contain Either All Correct or All Incorrect Poses}
\label{sec:ensemble_outcomes}

\begin{figure}[!t]
    \centering
    \includegraphics[width=0.85\textwidth]{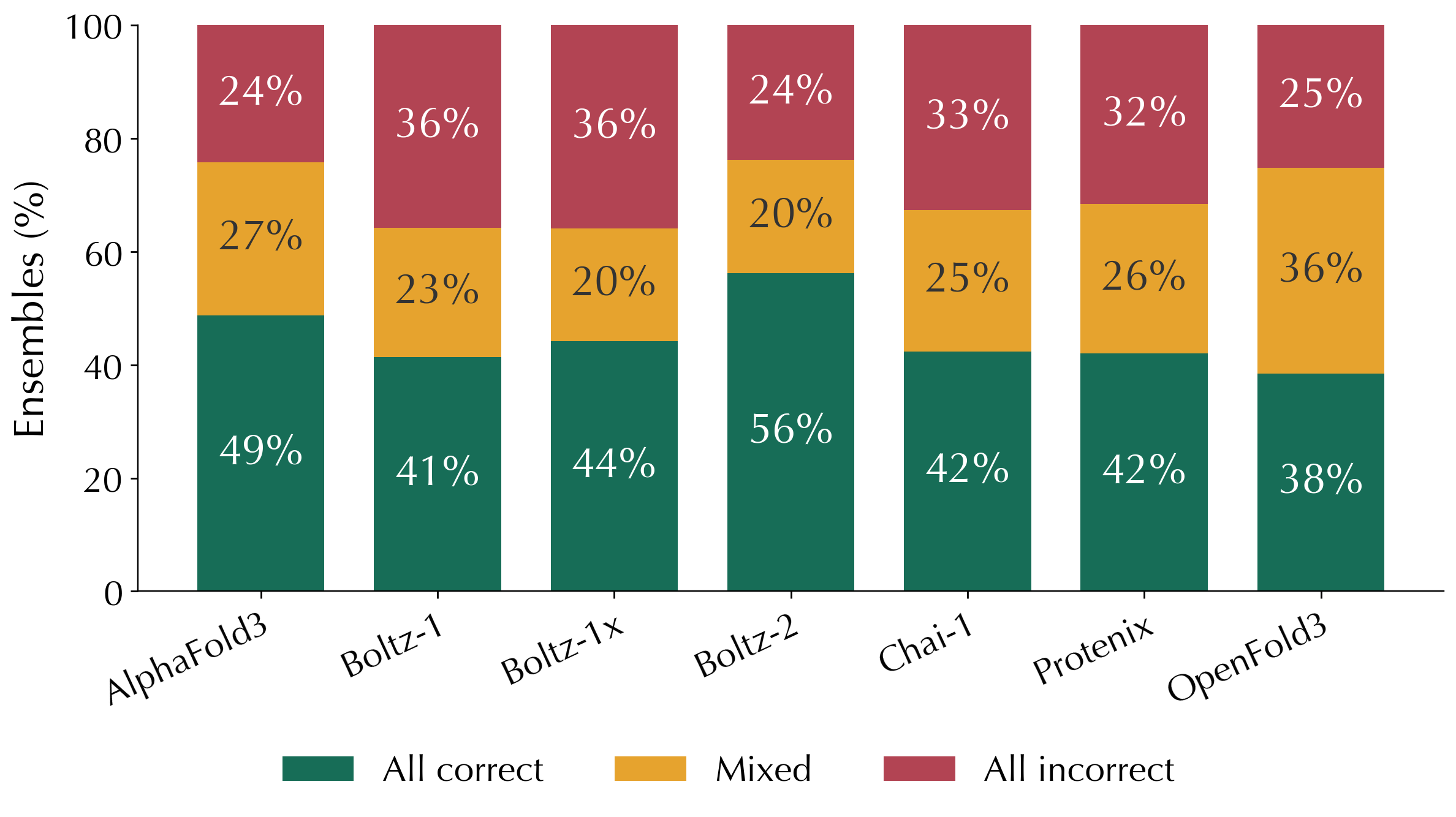}
    \caption{\textbf{Ensembles frequently contain either all correct or all incorrect poses.} Stacked bars show the percentage of ensembles in which all, some, or none of the 25 generated poses are structurally correct, separately for each co-folding model. In entirely correct or entirely incorrect ensembles, pose ranking cannot change the outcome.}
    \label{fig:ensemble_outcomes}
\end{figure}

To further investigate model ensembles, we grouped complex predictions according to whether all, some, or none of their poses were structurally correct (Figure~\ref{fig:ensemble_outcomes}). We found that most ensembles contained either entirely correct or entirely incorrect predictions, preventing ranking from changing the outcome regardless of which pose was selected. Averaged equally across the seven models, these two categories accounted for 74.6\% of ensembles. This limited the opportunity for confidence-based selection to improve on the random baseline. Additionally, among ensembles in which every pose passed the RMSD and lDDT-PLI thresholds, an average of 94.5\% contained at least one PoseBusters-valid pose, ranging from 91.3\% to 97.2\% across models (Figures~\ref{fig:si_valid_counts} and~\ref{fig:si_valid_allcorrect}). Most therefore contained at least one prediction satisfying all three criteria. Together, these results show that ensembles frequently contain either entirely correct or entirely incorrect predictions and that, when outcomes are mixed, confidence-based ranking offers only modest improvement over random sampling. This motivates an alternative approach for assessing predictive ensembles beyond existing confidence scores.

\FloatBarrier
\subsection{Structural Agreement Is Associated with Prediction Correctness}

The prevalence of entirely correct and entirely incorrect ensembles motivated us to examine whether these outcomes were reflected in the diversity of the generated poses, and therefore whether mode collapse is associated with correctness. We first compared diversity across whole ensembles, before examining whether the size and tightness of structural clusters were associated with the proportion of correct poses they contained.

\begin{figure}[!t]
    \centering
    \includegraphics[width=\textwidth]{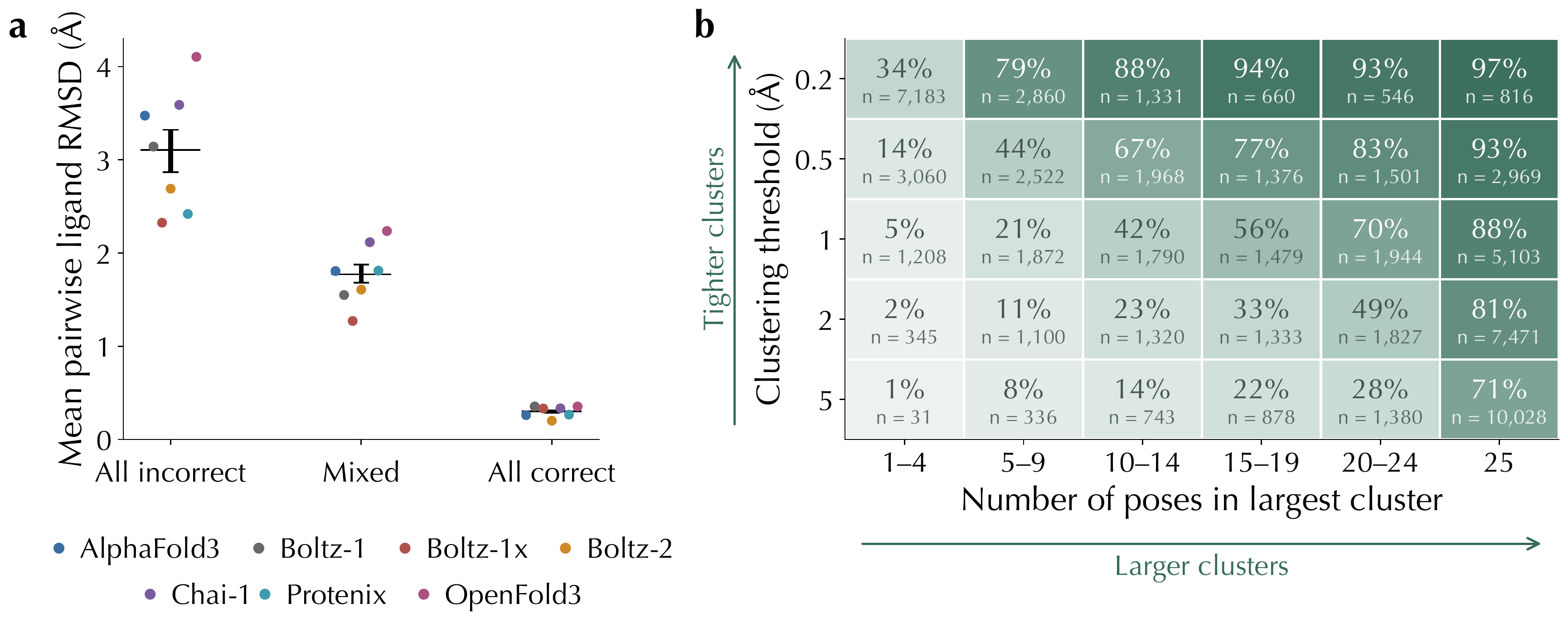}
    \caption{\textbf{Structural agreement is associated with prediction correctness.} (a) Mean pairwise ligand RMSD by ensemble outcome. Coloured points show model means with black marks showing the mean across models and 95\% bootstrap interval. (b) Percentage of correct poses in the largest cluster, averaged across models and grouped by cluster size and RMSD threshold used for clustering, with $n$ giving the number of ensembles pooled across models. Larger, more concentrated clusters generally contain more correct poses.}
    \label{fig:structural_modes}
\end{figure}

Entirely correct ensembles showed substantially less structural diversity than mixed or entirely incorrect ensembles, with mean pairwise ligand RMSDs of 0.30, 1.77 and 3.11~\AA, respectively, averaged across the seven models (Figures~\ref{fig:structural_modes}a,~\ref{fig:si_diversity} and~\ref{fig:si_diversity_full}). While entirely correct ensembles must, by definition, be composed of a low diversity ensemble to remain below $2$~\AA\ RMSD to the ground truth, there is no equivalent constraint on entirely incorrect ensembles.

We then further examined how this relationship extended to structural groups within each ensemble. Poses were grouped into clusters in which every pair satisfied a specified ligand RMSD cutoff. After clustering, larger and more tightly agreeing clusters generally contained a higher proportion of correct poses (Figure~\ref{fig:structural_modes}b). At a cutoff of $0.5$~\AA, the mean correct fraction within the largest mode increased from 14\% for modes containing 1 to 4 poses to 93\% for those containing all 25 poses in the ensemble. Additionally, for clusters containing 15 to 19 poses, the proportion of correct poses increased from 22\% when using a clustering threshold of $5$~\AA, to 94\% when using $0.2$~\AA. These trends held across all models, as shown by model-specific results in Figure~\ref{fig:si_mode_size}.

Based on these findings, we combined mode tightness and ensemble support into CORE Score (Section~\ref{sec:quantifying}). We next tested whether CORE Score could indicate whether the top-ranked prediction is likely to be correct, using only the generated pose ensemble in a setting reflecting prospective use.

\subsection{CORE Score Improves Prediction Discrimination}
\label{sec:prospective}

CORE Score was calculated using all 25 generated poses, including physically invalid predictions, as described in Section~\ref{sec:quantifying}. However, in prospective use, physically invalid poses can be identified and discarded before selecting a prediction. To reflect this, we excluded PoseBusters-invalid poses from pose selection, selecting the top-ranked prediction from the remaining valid poses. A prediction was therefore defined as correct if it had an RMSD below $2$~\AA, lDDT-PLI above $0.8$, and was PoseBusters-valid.

\begin{figure}[!t]
    \centering
    \includegraphics[width=\textwidth]{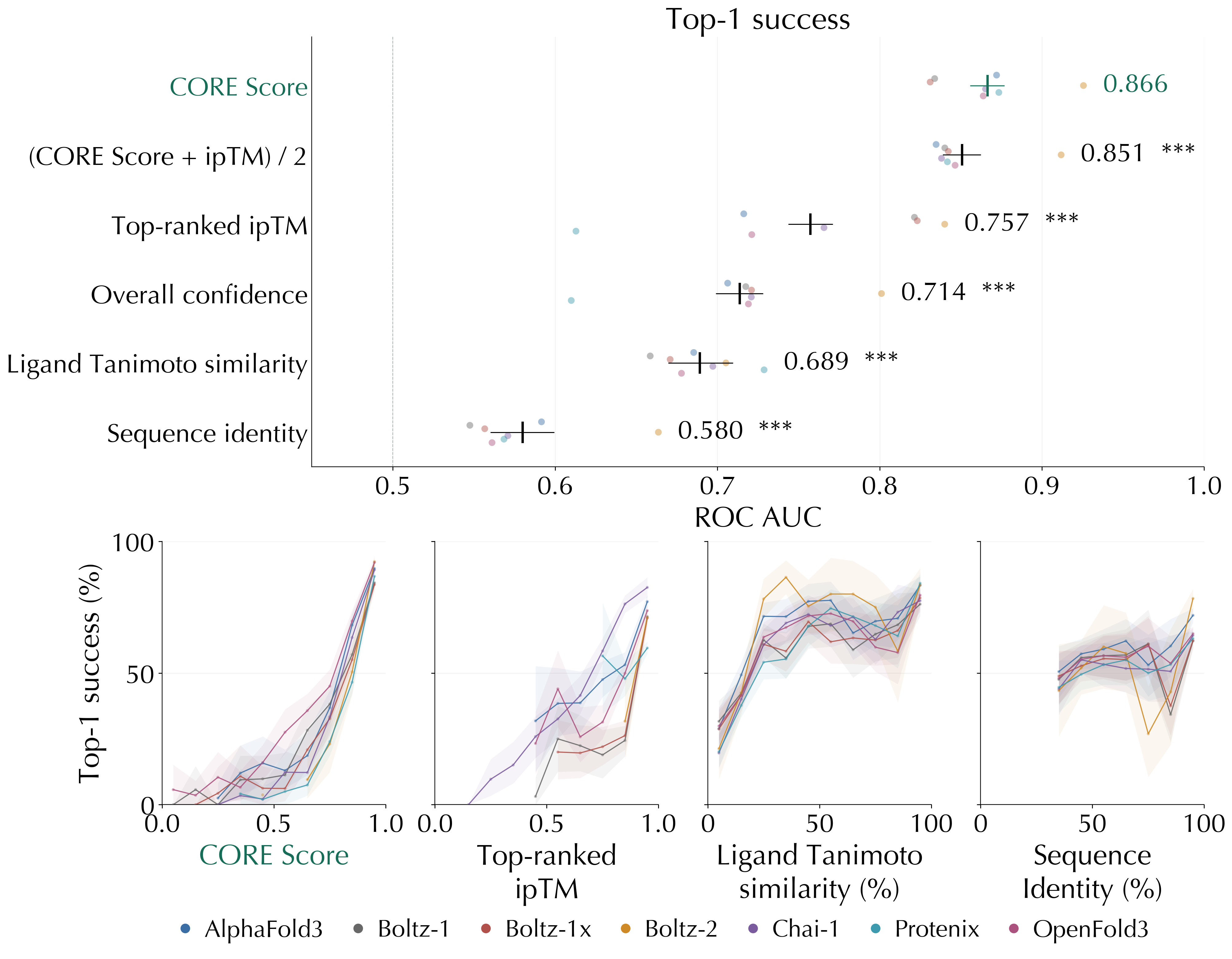}
    \caption{\textbf{Structural agreement discriminates prediction outcomes and identifies groups with different success rates.} Top: ROC AUCs for top-1 success. Coloured points identify models. Black marks show equal-model means with 95\% confidence intervals, and asterisks mark differences from CORE Score in equal-model mean ROC AUC ($^{***}$two-sided $p \leq 0.001$), each from the same paired PDB-entry bootstrap. Bottom: top-1 success across CORE, ipTM, ligand Tanimoto, and sequence identity bins. Bands show pointwise 95\% PDB-entry bootstrap intervals. Bins with fewer than 20 ensembles are omitted. All comparisons use the matched PB-valid candidate pools. All displayed measures evaluate the highest-ipTM pose, except overall confidence, which evaluates its own top-ranked pose. Oracle success and pairwise RMSD comparisons are provided in Figures~\ref{fig:si_oracle_auc} and~\ref{fig:si_valid_alternatives_auc}.}
    \label{fig:core_comparison}
\end{figure}

CORE Score achieved an average AUC of 0.866 across the 7 models for top-1 success, outperforming ligand-protein chain-pair ipTM (0.757) and overall confidence (0.714). Tanimoto similarity and sequence identity showed poorer ability to distinguish correct predictions, with sequence identity being the least informative (Figure~\ref{fig:core_comparison}). Combining CORE Score and ipTM by taking their mean gave an average AUC of 0.851, lower than CORE Score alone, although performance was more uniform across models. The combination improved top-1 AUC for Boltz-1 by 0.006 and for Boltz-1x by 0.011, consistent with ipTM performing comparatively well for these models. CORE Score also achieved the highest AUC for oracle success, at 0.838 compared with 0.759 for ipTM (Figure~\ref{fig:si_oracle_auc}). Minimum and mean pairwise ligand RMSD provided strong, simpler descriptors of diversity, with AUCs of 0.850 and 0.848 respectively for top-1 success (Figures~\ref{fig:si_valid_alternatives_auc} and~\ref{fig:si_core_vs_min_rmsd}). CORE Score retained its advantage when invalid poses were also included (Figures~\ref{fig:si_allposes_auc} and~\ref{fig:si_allposes_curves}) and when correctness was defined using stricter RMSD thresholds (Tables~\ref{tab:si_threshold} and~\ref{tab:si_threshold_grid}).

We next examined how prediction success varied across the range of each measure (Figure~\ref{fig:core_comparison}, bottom, and Figure~\ref{fig:si_valid_alternatives_curves} for the pairwise RMSD baseline comparisons). CORE Score showed a more gradual increase in the proportion of correct top-ranked predictions, with similar trends across all seven models. In comparison, ipTM showed greater variation between models, while the similarity measures showed less consistent increases in success across their ranges. We also compared the most and least favourable 10\% of ensembles for each score within each model. Averaged across models, 96.1\% of top predictions were correct among the 10\% with the highest CORE Scores, compared with 92.1\% for the highest ipTM and 80.8\% for the highest ligand Tanimoto similarity. Among the 10\% with the lowest CORE Scores, 5.8\% were correct, compared with 27.3\% and 24.9\% for the lowest ipTM and Tanimoto similarity, respectively. This indicates that CORE Score more clearly separates ensembles with high and low likelihoods of a correct top-ranked prediction than the other metrics compared here, with the comparison against ipTM repeated at every other cutoff in Figure~\ref{fig:si_retention}.

\subsection{CORE Score Provides Stronger Discrimination than Structural Similarity}

Similarity to training examples has been shown to provide a strong indication of prediction accuracy. However, assessing this similarity using structure-based metrics requires experimental structural information, which is unavailable in prospective applications. For instance, Runs N' Poses assesses this similarity using SuCOS shape $\times$ pocket similarity, a structure-based measure incorporating the ligand binding pose and its surrounding pocket \cite{skrinjar_have_2025, leung_sucos_2019}. In contrast, CORE Score can be calculated from generated structures alone, making it more applicable to prospective applications where the ground truth pose is unavailable. We therefore compared CORE Score with this metric for top-1 success and oracle success using ROC AUC.

\begin{figure}[!t]
    \centering
    \includegraphics[width=1.0\textwidth]{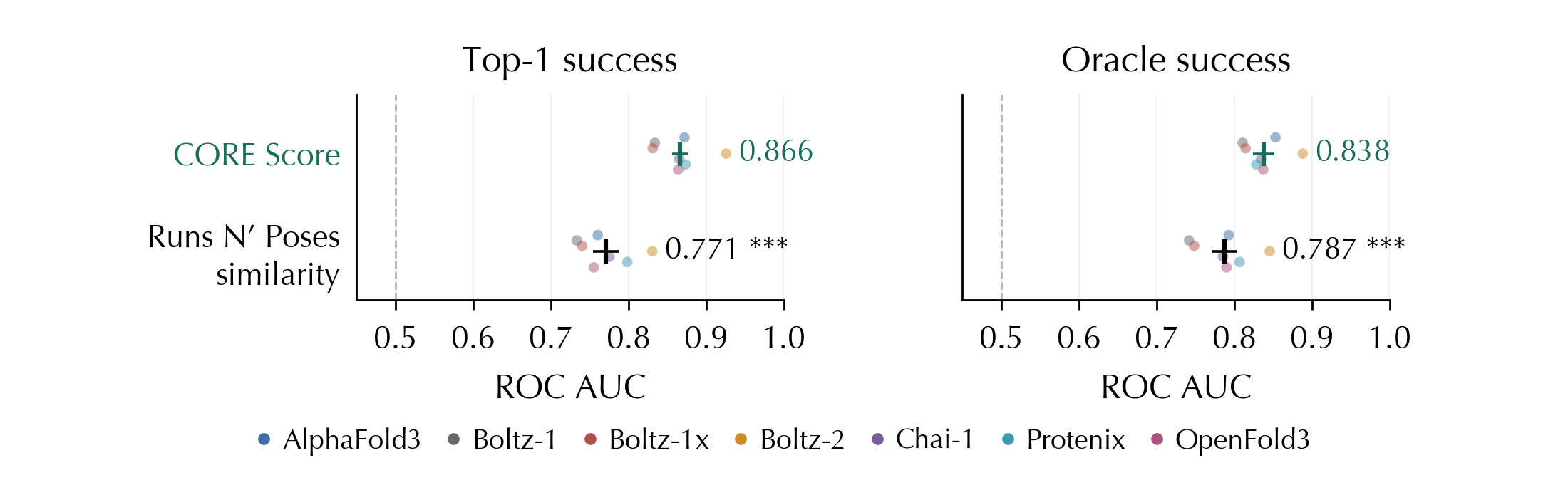}
    \caption{\textbf{CORE Score provides stronger discrimination than structural similarity.} ROC AUCs for top-1 success (left) and oracle success (right), comparing CORE Score with the Runs N' Poses similarity metric. Coloured points identify models. Black marks show equal-model means with 95\% confidence intervals, and asterisks mark differences from CORE Score in equal-model mean ROC AUC ($^{***}$two-sided $p \leq 0.001$), each from the same paired PDB-entry bootstrap.}
    \label{fig:s3_similarity_auc}
\end{figure}

CORE Score achieved higher discrimination than the Runs N' Poses metric across all seven models for both outcomes (Figure~\ref{fig:s3_similarity_auc}). Averaged equally across models, ROC AUCs were 0.866 versus 0.771 for top-1 success and 0.838 versus 0.787 for oracle success. CORE Score therefore provided stronger discrimination of both outcomes, without requiring the experimental ligand pose.

\begin{figure}[!t]
    \centering
    \includegraphics[width=0.7\textwidth]{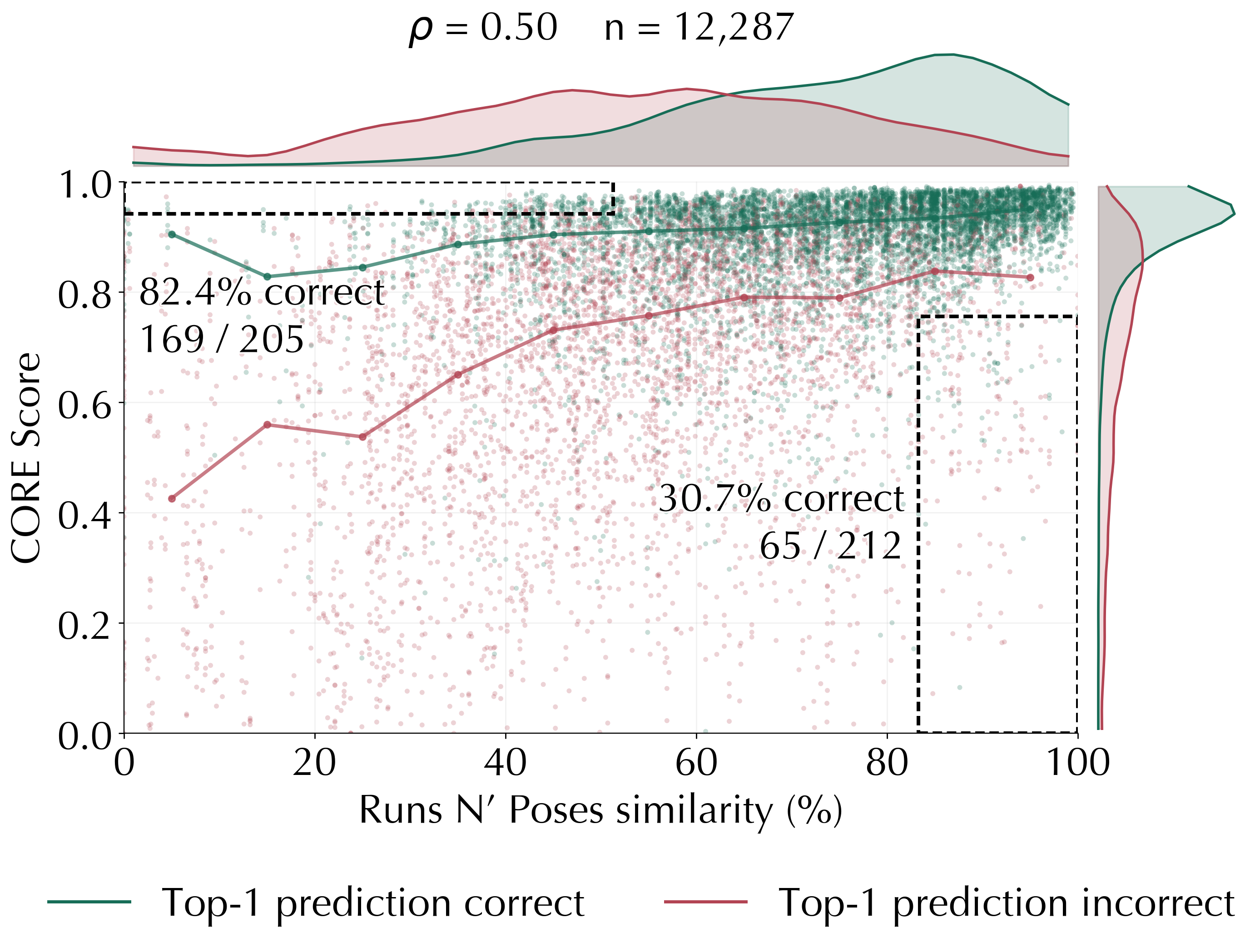}
    \caption{\textbf{CORE Score and structural similarity can give different indications of correctness.} Points show ensembles coloured by top-1 success. Dashed boxes highlight regions where one measure is in its upper quartile and the other in its lower quartile, with prediction success rates shown.}
    \label{fig:s3_similarity_discrimination}
\end{figure}

CORE Score was positively correlated with the Runs N' Poses metric across all seven models, with Spearman correlations ranging from 0.45 to 0.67 (Figure~\ref{fig:si_s3_correlation_models}) and a pooled correlation of 0.50. However, the two measures did not always give the same indication of prediction correctness (Figure~\ref{fig:s3_similarity_discrimination}). We therefore examined two groups where their indications differed. Each group was defined by the lower and upper quartiles of each measure across all 12,287 ensembles. In the group favoured by CORE Score (CORE Score $\geq0.942$, and similarity $\leq51.3\%$), 82.4\% of top predictions were correct (169 of 205). Conversely, in the group favoured by structural similarity (similarity $\geq83.3\%$, and CORE Score $\leq0.755$), only 30.7\% were correct (65 of 212). These exploratory comparisons support CORE Score as a stronger indicator of top-1 success than structural similarity in the two groups where the measures give opposing indications.

\begin{figure}[!t]
    \centering
    \includegraphics[width=\textwidth]{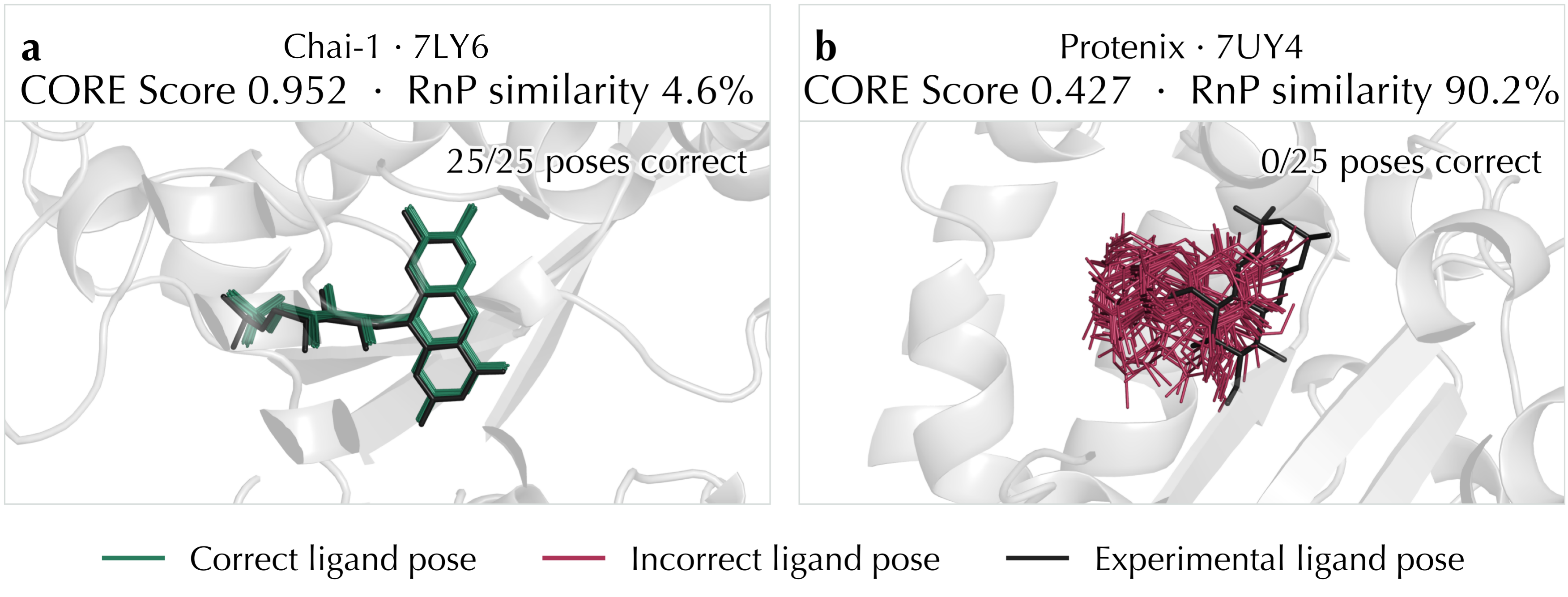}
    \caption{Predicted ligand poses for (a) 7LY6 from Chai-1 and (b) 7UY4 from Protenix, overlaid on the experimental ligand (black).}
    \label{fig:s3_examples}
\end{figure}

Two examples illustrate these opposing cases (Figure~\ref{fig:s3_examples}). For 7LY6, the Runs N' Poses similarity was only 4.6\%, yet every model produced 25 of 25 poses with RMSD below 2~\AA{} and lDDT-PLI above 0.8, with a mean CORE Score across models of 0.928. The example shows these results for Chai-1, which had a CORE Score of 0.952. Conversely, the Runs N' Poses similarity for 7UY4 was high at 90.2\%, yet four of the six models produced no correct poses. Protenix, for example, produced a diffuse ensemble of incorrect poses, giving a CORE Score of 0.427. In both cases, CORE Score indicated the outcome correctly where structural similarity did not. Results for both targets across all models are shown in Figure~\ref{fig:si_examples_all_models}.

Together, these results show that CORE Score is associated with structural similarity to training examples, but provides a stronger indication of prediction correctness when the two measures disagree. Additionally, CORE Score is both model-specific and calculable without the ground truth, making it better suited to prospective applications.

\section{Discussion}

In this work, we first assess the ability of co-folding models to rank poses and to distinguish correct from incorrect predictions, showing that learned confidence scores offer limited improvement over random selection within mixed ensembles and only modest discrimination of prediction correctness across complexes. We then show that structural agreement among repeated predictions provides a useful indicator of prediction correctness across seven co-folding models. After filtering physically invalid poses, CORE Score achieves mean ROC AUCs of 0.866 for top-1 success and 0.838 for oracle success, compared with 0.757 and 0.759 for ipTM. Averaged across models, top-1 success reaches 96.1\% among the highest-scoring tenth of ensembles, compared with 5.8\% among the lowest. Simpler summaries of structural agreement, such as minimum, median, and mean pairwise RMSD, also perform strongly, with CORE Score modestly ahead for both top-1 and oracle success. Finally, CORE Score achieves higher mean discrimination than Runs N' Poses similarity, which requires the experimental ligand pose. These results support structural agreement as a practical reliability signal, without implying that a more complex concentration formula is always necessary.

Ensembles in which all 25 poses were correct were far more concentrated than those in which none were correct. This held for every model, indicating that convergence on a single solution was strongly associated with correctness. While convergence is necessary for every pose in an ensemble to be correct, it is not sufficient. An ensemble may equally collapse onto a single incorrect solution. We find structural agreement to strongly indicate correct predictions. This association was not limited to complexes highly similar to the training data (Figure~\ref{fig:si_curves_by_similarity}). Among complexes with Runs N' Poses similarity below 50\%, ensembles in which all poses were correct remained tightly concentrated (median pairwise RMSD 0.24~\AA{}, compared with 1.78~\AA{} when none were correct). Similarly, high similarity did not guarantee success, with 367 ensembles of at least 80\% similarity containing no correct poses.

One possible explanation for the advantage of CORE Score over metrics used to assess similarity to the training set is that assessing accuracy through similarity to training examples requires choosing which aspects of resemblance are relevant to model performance. Sequence, chemical, and structural similarity provide principled measures of familiarity, and the performance of the Runs N' Poses metric demonstrates their value. However, the model may learn patterns in the training data that we are unaware of and, therefore, these similarity measures chosen by us do not capture. Complexes that appear similarly familiar under a chosen metric may therefore differ in how reliably the model can predict them.

The usefulness of CORE Score rests on how current models behave when sampled repeatedly. Convergence on a single prediction indicates how confident the model is about that prediction, and therefore how reliable the prediction itself is likely to be. However, this interpretation depends on our current definition of success, which measures closeness to a single experimental pose. A crystal structure captures only one state of a dynamic system, and a ligand may occupy an ensemble of low-energy states in its binding site. We may prefer a model that samples from the Boltzmann distribution of bound states, such that diversity among generated poses reflects alternative low-energy states rather than uncertainty in the prediction. This leaves an open question. While ensemble diversity is currently informative about a model’s predictive certainty, it is unclear whether near-identical sampling is a desirable behaviour, given that the true bound state is better described by an ensemble of low-energy conformations.

Further work could examine how changes to the sampling procedure could affect ensemble diversity and the ability of CORE Score to assess prediction reliability. Additionally, our analysis has focused on the correctness of protein-ligand structures. Whether structural agreement can similarly inform the reliability of other structural predictions, such as antibody-antigen structures, protein-protein complexes, or protein-nucleic acid complexes, remains to be evaluated.

\bigskip

\subsection*{Funding}

J. Henry Broster acknowledges funding from the Engineering and Physical Sciences Research Council (EPSRC) under grant number EP/S024093/1 and from the Cambridge Crystallographic Data Centre through the SABS:R3 doctoral training programme.

\subsection*{Data availability}

Predicted structures, confidence scores, reference RMSD and lDDT-PLI measurements, PoseBusters checks, and training-set similarity measures for AlphaFold3, Boltz-1, Boltz-1x, Boltz-2, Chai-1, and Protenix are available from the Runs N' Poses benchmark \cite{skrinjar_have_2025}. OpenFold3 predictions were obtained through its GitHub-linked AWS distribution \cite{openfold3-preview}.

\subsection*{Code availability}

Code to compute CORE Score is available at \url{https://github.com/oxpig/core-score}.

\clearpage
\bibliography{sn-bibliography}
\clearpage
\raggedbottom
\section*{Supplementary Information}
\setcounter{figure}{0}
\renewcommand{\thefigure}{S\arabic{figure}}
\setcounter{table}{0}
\renewcommand{\thetable}{S\arabic{table}}

\subsection*{S1. Multiple ligand copies}
The following examples show multiple ligand copies, with CORE Scores calculated separately for each copy or site using all 25 predictions before PB filtering.

\begin{figure}[H]
\centering
\includegraphics[width=\textwidth]{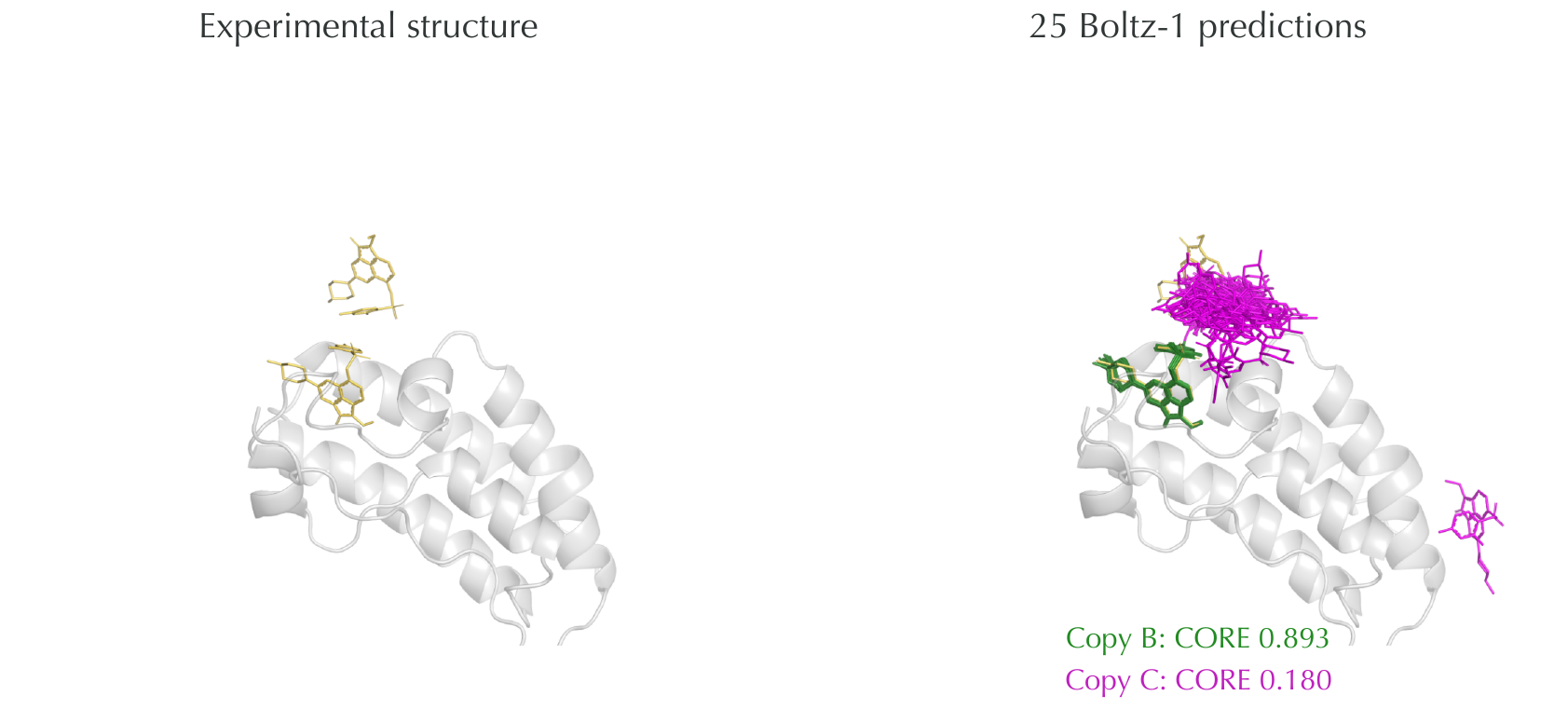}
\caption{\textbf{Experimental and predicted ligand copies in 7WL4.} Left: experimental structure, with ligands in gold. Right: 25 aligned Boltz-1 predictions, with copies B and C in green and magenta. Their CORE Scores are 0.893 and 0.180, respectively.}
\label{fig:si_7wl4}
\end{figure}
\begin{figure}[H]
\centering
\includegraphics[width=\textwidth]{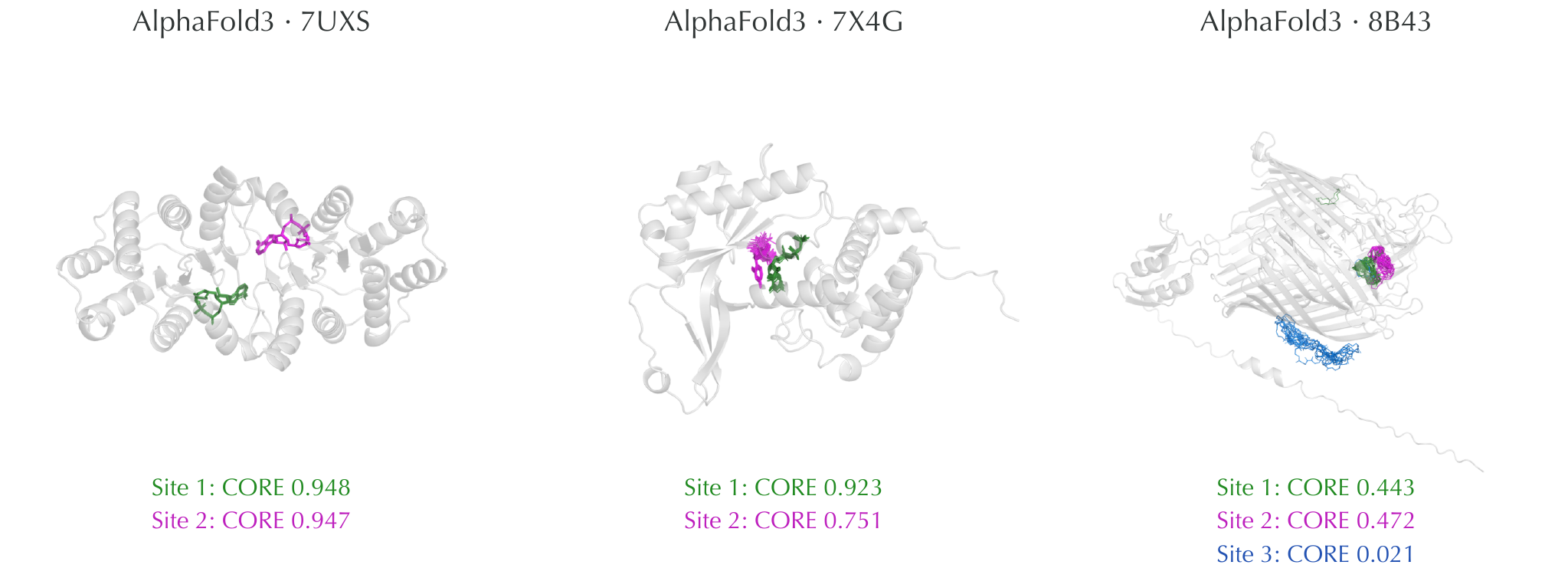}
\caption{\textbf{CORE Scores for individual ligand sites.} Three AlphaFold3 examples, each showing 25 predictions after protein alignment. Colours distinguish ligand sites, with a CORE Score shown for each site.}
\label{fig:si_multi_examples}
\end{figure}

\newpage

\subsection*{S2. Data coverage}
Before similarity filtering, the retained single-copy dataset comprised 13,396 ensembles, each containing 25 original poses and complete paired reference RMSD/lDDT-PLI labels. Similarity filtering retained 12,990 ensembles, as shown in Table~\ref{tab:si_coverage}.

\begin{table}[htbp]
\centering
\small
\caption{Model-specific analysis-group counts. Proper denotes groups with at least one released prediction record having \texttt{ligand\_is\_proper=True}. For OpenFold3, proper-ligand eligibility was determined from reference annotations. Multiple-copy denotes subsequent exclusions involving multiple copies of the evaluated ligand. Missing similarity denotes exclusions lacking protein sequence identity, ligand Tanimoto similarity, or Runs N' Poses similarity annotations. Each retained group contains 25 generated poses.}
\label{tab:si_coverage}
\begin{tabular}{lrrrrr}
\hline
Model & Available & Proper & Multiple-copy & \shortstack{Missing similarity} & Retained \\
\hline
AlphaFold3 & 2,679 & 2,679 & 556 & 47  & 2,076 \\
Boltz-1    & 2,400 & 2,356 & 352 & 47  & 1,957 \\
Boltz-1x   & 2,304 & 2,304 & 346 & 44  & 1,914 \\
Boltz-2    & 953   & 926   & 143 & 20  & 763   \\
Chai-1     & 2,543 & 2,543 & 457 & 46  & 2,040 \\
Protenix   & 2,594 & 2,579 & 482 & 46  & 2,051 \\
OpenFold3  & 2,927 & 2,927 & 582 & 156 & 2,189 \\
\hline
Total & 16,400 & 16,314 & 2,918 & 406 & 12,990 \\
\hline
\end{tabular}
\end{table}

\FloatBarrier
\newpage

\subsection*{S3. Availability of physically valid poses}
\begin{figure}[H]
\centering
\includegraphics[width=\textwidth]{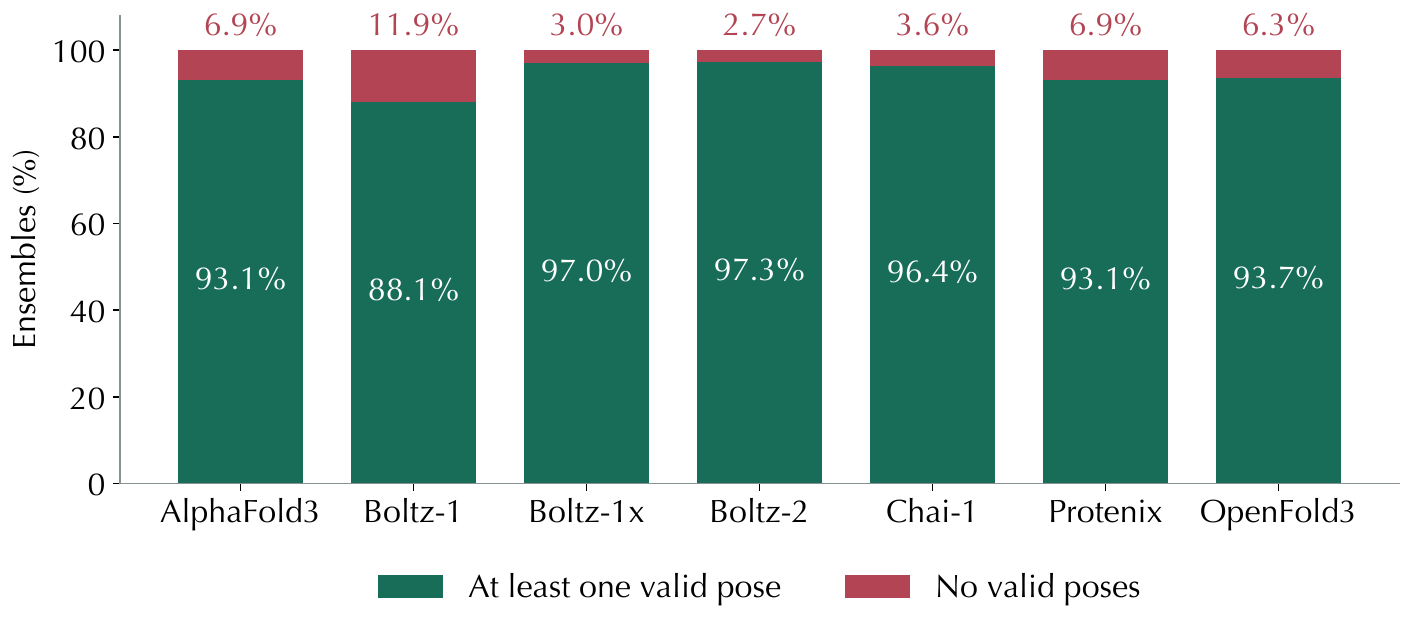}
\caption{\textbf{Availability of physically valid poses.} Bars show the percentage of ensembles containing at least one PoseBusters-valid pose or no valid poses.}
\label{fig:si_valid_counts}
\end{figure}

\begin{figure}[H]
\centering
\includegraphics[width=\textwidth]{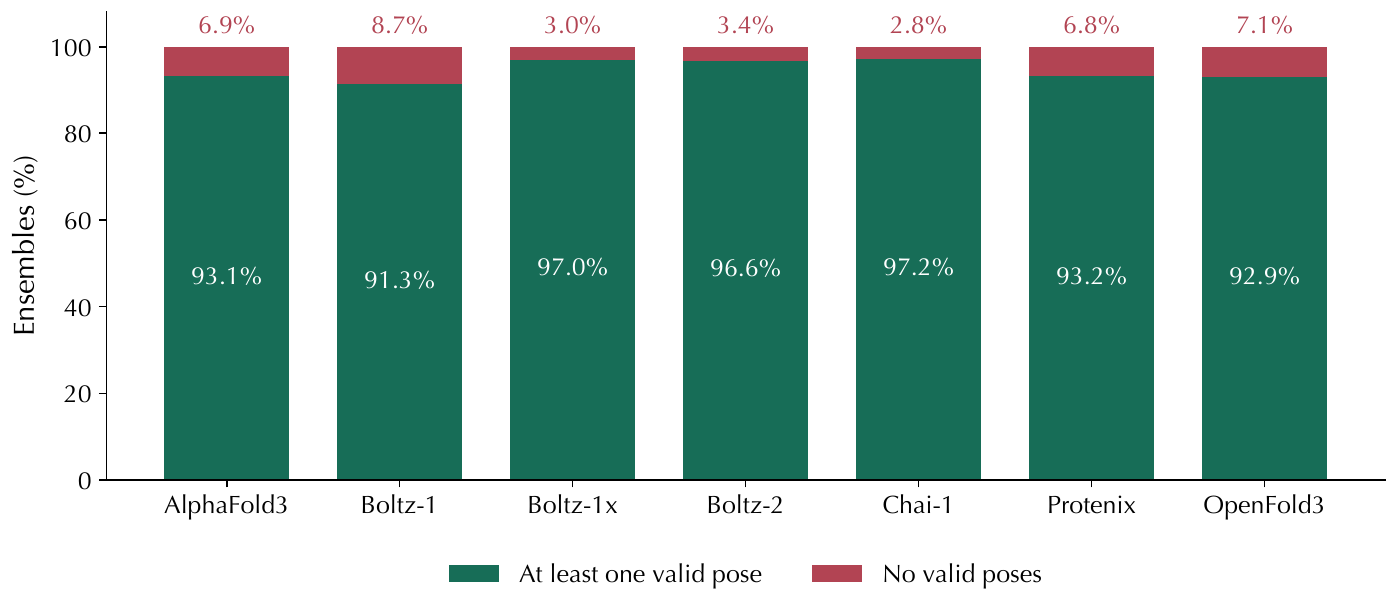}
\caption{\textbf{Availability of physically valid poses in fully correct ensembles.} Bars show the percentage of ensembles containing at least one PoseBusters-valid pose or no valid poses, among ensembles in which every pose had RMSD below $2$~\AA\ and lDDT-PLI above $0.8$.}
\label{fig:si_valid_allcorrect}
\end{figure}

Figure~\ref{fig:si_valid_counts} shows whether each ensemble contains at least one PoseBusters-valid pose. Figure~\ref{fig:si_valid_allcorrect} gives the same quantity among ensembles in which every pose had RMSD below $2$~\AA\ and lDDT-PLI above $0.8$.

\FloatBarrier

\newpage

\subsection*{S4. Overall confidence comparisons}
These comparisons use the same 13,396 single-copy ensembles as the main ipTM analysis, retaining all 25 poses without PB filtering. Structural correctness requires RMSD below $2$~\AA\ and lDDT-PLI above $0.8$. Averaging equally across the seven models, ipTM-based selection succeeds in 60.4\% of ensembles, compared with 60.0\% for overall confidence and 58.6\% for random selection. Within mixed ensembles, the corresponding rates are 61.4\%, 60.3\% and 54.3\%.

\begin{figure}[H]
\centering
\includegraphics[width=\textwidth]{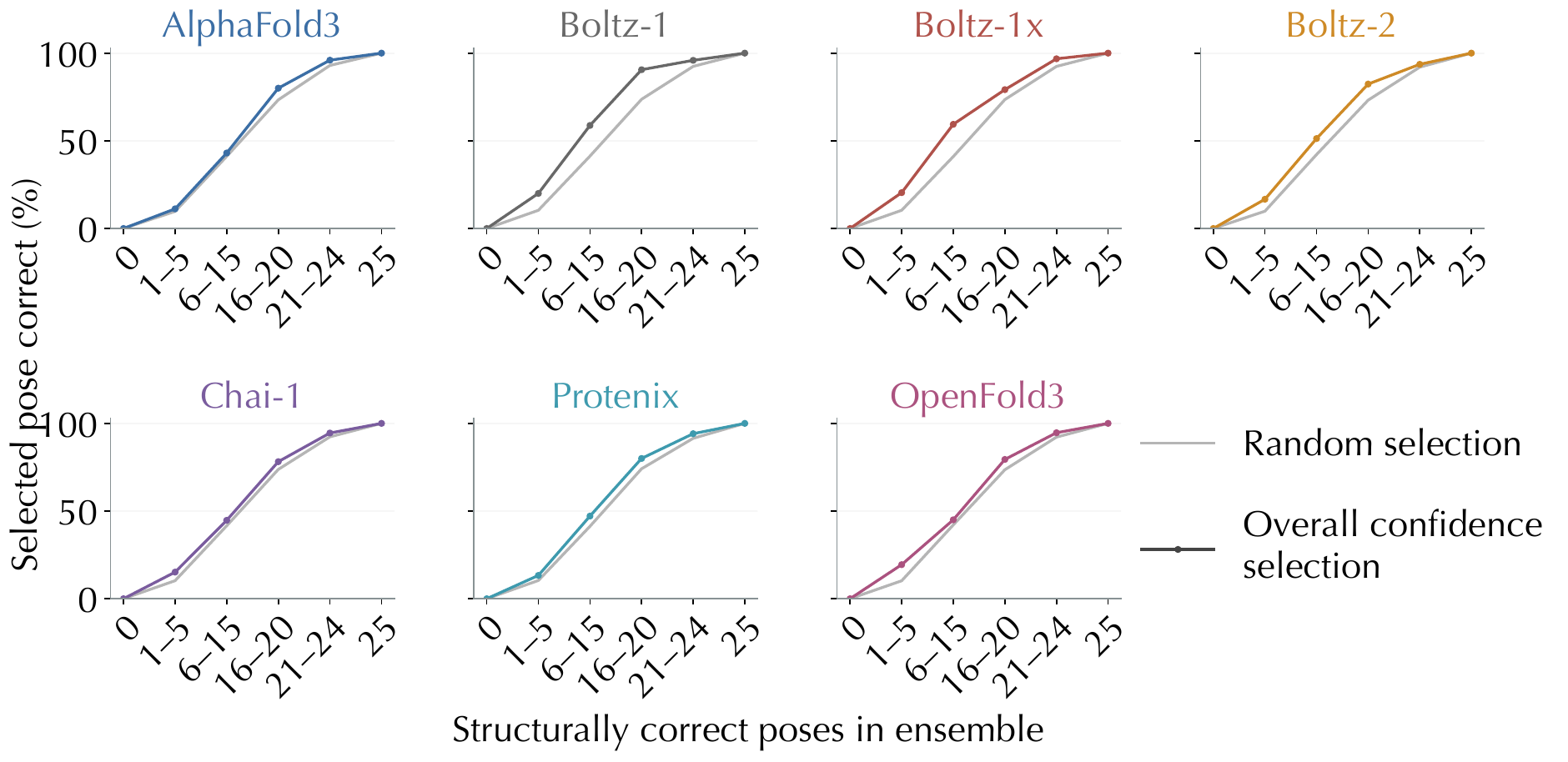}
\caption{\textbf{Pose selection using overall confidence.} The highest overall confidence pose is compared with uniform random selection. All original poses are retained; correctness is structural only.}
\label{fig:si_sel_native}
\end{figure}
\begin{figure}[H]
\centering
\includegraphics[width=\textwidth]{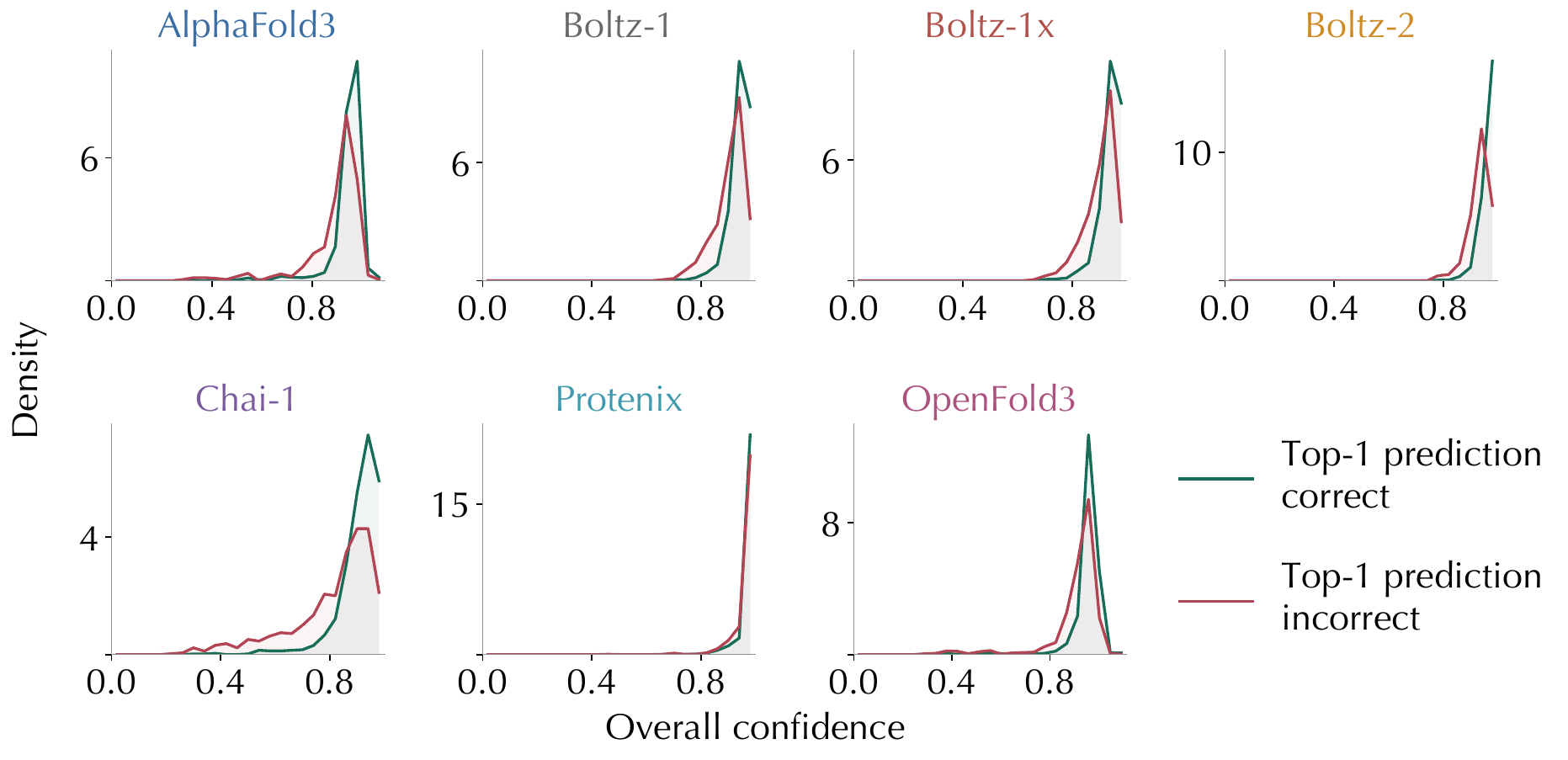}
\caption{\textbf{Overall confidence distributions by top-1 success.} Each ensemble contributes the overall confidence of its highest-overall confidence prediction. Correct and incorrect distributions are normalised separately. All original poses are retained, and PoseBusters-validity is not required for structural correctness.}
\label{fig:si_dist_native}
\end{figure}
\begin{figure}[H]
\centering
\includegraphics[width=\textwidth]{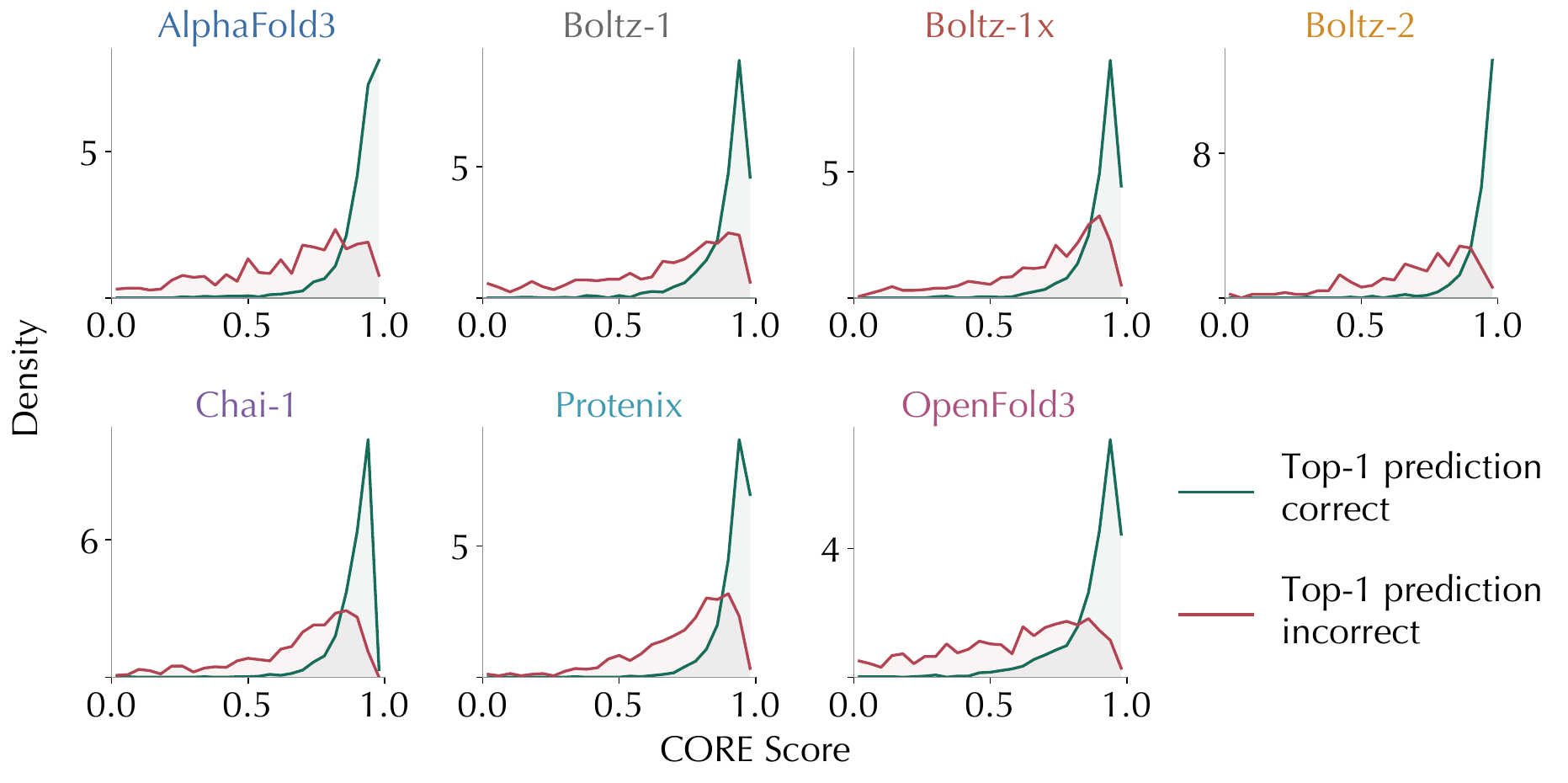}
\caption{\textbf{CORE Score distributions by top-1 success before validity filtering.} CORE Score is calculated using all 25 poses and grouped by whether the highest-ipTM prediction is structurally correct. Green and red distributions are normalised separately; they do not represent outcome prevalence. These are the same 13,396 ensembles and highest-ipTM selected poses as in the main ipTM comparison; overall confidence in the preceding supplementary figures selects its own pose.}
\label{fig:si_core_dist}
\end{figure}

\FloatBarrier

\subsection*{S5. Pairwise RMSD Calculations}
\phantomsection\makeatletter\def\@currentlabel{S5}\makeatother
\label{sec:si_methods}

Pairwise ligand RMSDs were calculated after binding-site superposition using OpenStructure 2.8.0 \cite{ost_2026}, with \texttt{compare-ligand-structures}, \texttt{-rmsd}, \texttt{-full-bs-search}, \texttt{-full-results}, and \texttt{-fault-tolerant}. Comparisons used generated structures only, with a default binding-site radius of $4$~\AA. Three ensembles required $6$~\AA\ fallbacks; excluding them changed evaluated AUCs by less than 0.001. Reference RMSD and lDDT-PLI measurements described the same predicted copy and experimental ligand. Complete-linkage clustering used symmetric pairwise RMSD matrices. Correct fractions were averaged across tied largest clusters, then within each model and equally across models.

\newpage

\subsection*{S6. Baseline comparisons}
Averaged equally across models, ROC AUCs for top-1 success were 0.850 for minimum, 0.849 for median, and 0.848 for mean pairwise RMSD, compared with 0.866 for CORE Score. For oracle success they were 0.829, 0.817, and 0.812 respectively, compared with 0.838 for CORE Score.
\begin{figure}[H]
\centering
\includegraphics[width=\textwidth]{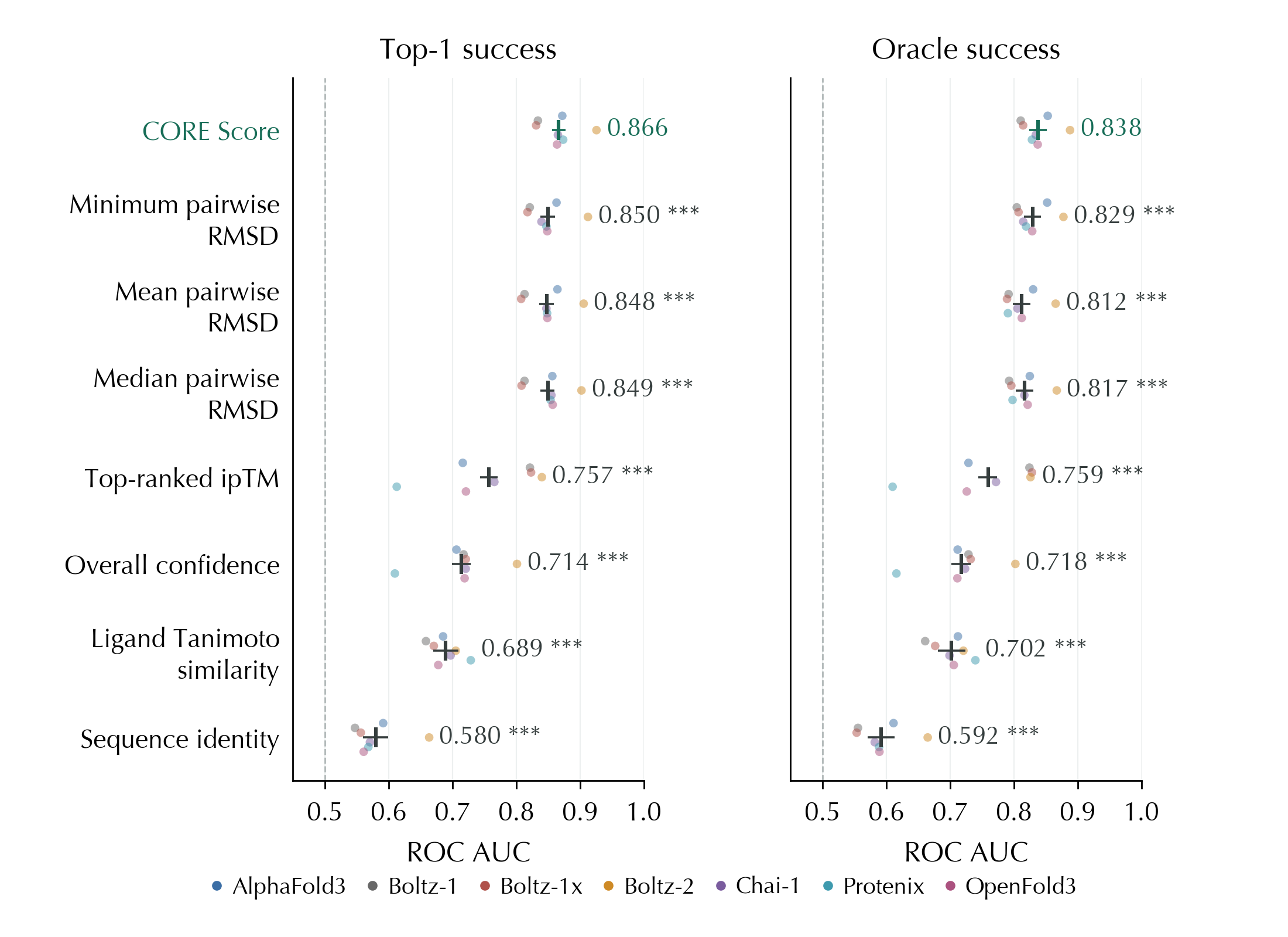}
\caption{\textbf{CORE Score and minimum pairwise RMSD compared with confidence and similarity.} Candidates are PB-valid, and joint success requires RMSD below $2$~\AA, lDDT-PLI above $0.8$, and PoseBusters-validity. Coloured points identify models. Black marks show equal-model means with 95\% confidence intervals, and asterisks mark differences from CORE Score in equal-model mean ROC AUC ($^{***}$two-sided $p \leq 0.001$), each from the same paired PDB-entry bootstrap.}
\label{fig:si_valid_alternatives_auc}
\end{figure}
\begin{figure}[H]
\centering
\includegraphics[width=\textwidth]{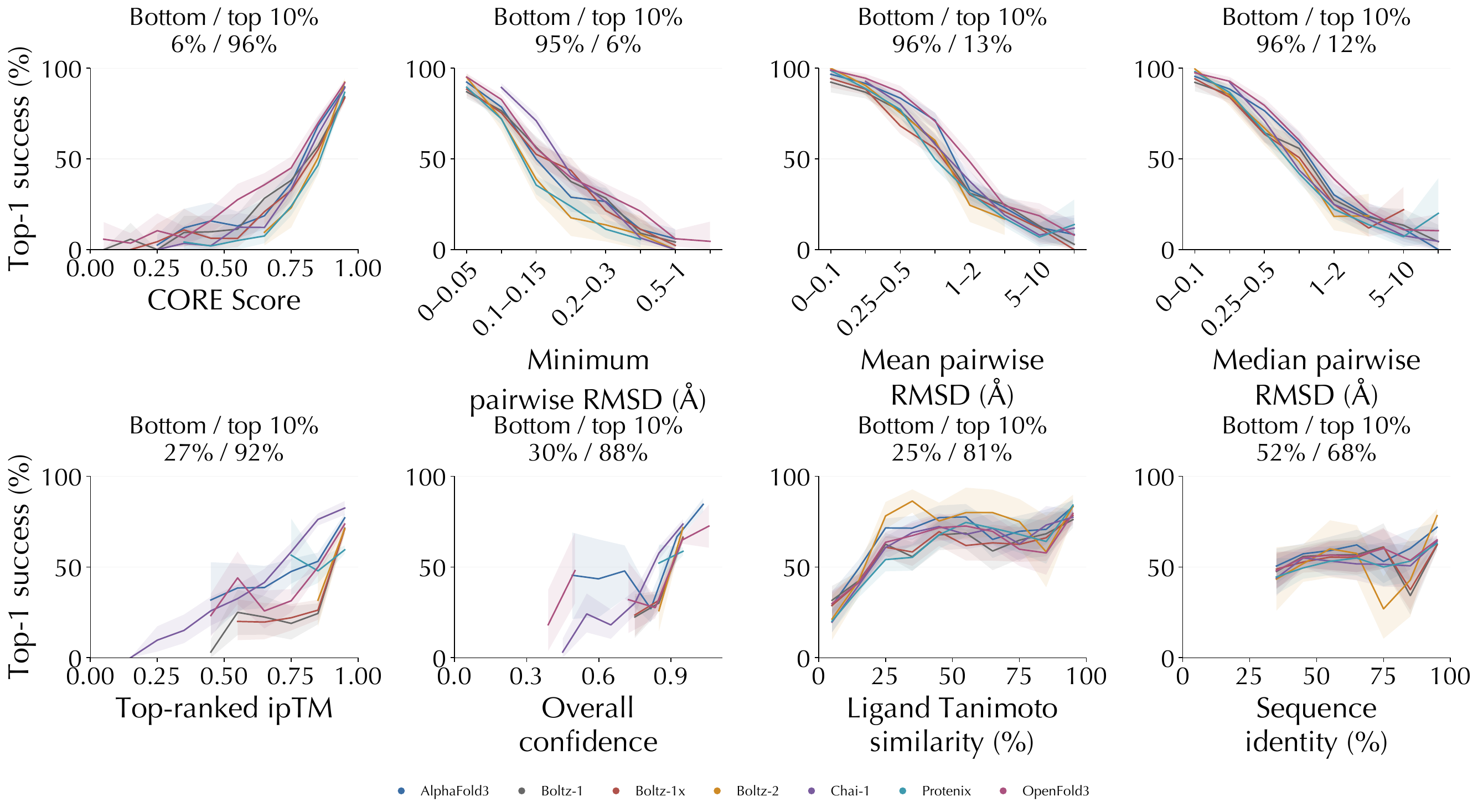}
\caption{\textbf{Score-success curves including minimum pairwise RMSD.} All eight measures use the same matched PB-valid cohorts. Bands show pointwise 95\% PDB-entry bootstrap intervals. The pairwise RMSD axes show raw distances in \AA; smaller values indicate greater agreement.}
\label{fig:si_valid_alternatives_curves}
\end{figure}
CORE Score and minimum pairwise RMSD were strongly correlated (Spearman $\rho = 0.92$), so we compared them where they disagreed. Both measures were converted to ranks within each model, and disagreement was taken as the difference between the two ranks. In the 10\% of ensembles that CORE Score ranked furthest above minimum pairwise RMSD, 71.4\% of top predictions were correct (1,230 ensembles), compared with 59.4\% in the 10\% that minimum pairwise RMSD ranked furthest above CORE Score (1,230 ensembles) and 59.5\% where the two agreed.
\begin{figure}[H]
\centering
\includegraphics[width=\textwidth]{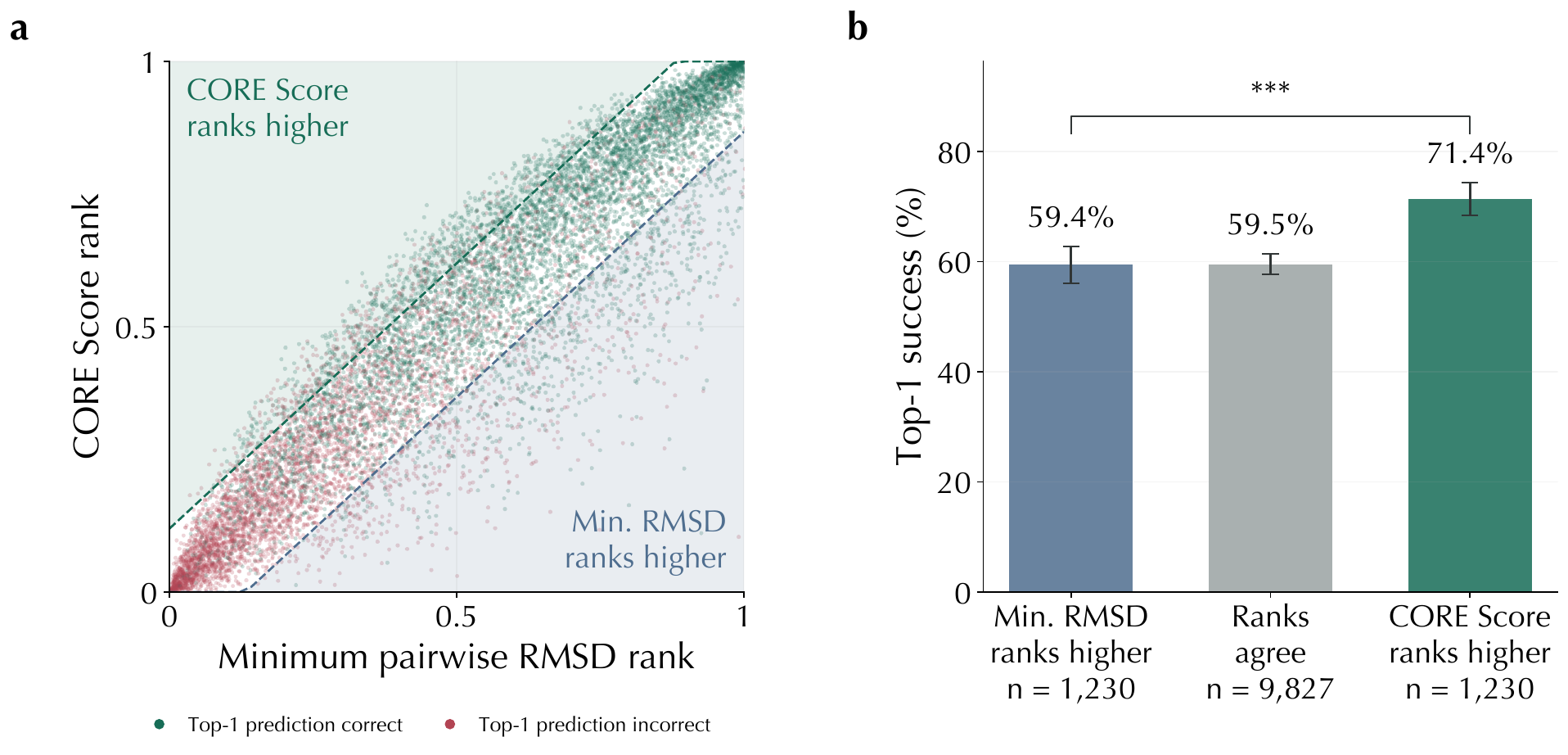}
\caption{\textbf{CORE Score against minimum pairwise RMSD.} (a) Ranks of the two measures within each model for PB-valid ensembles, coloured by top-1 success. Shaded regions mark the 10\% of ensembles each measure ranks furthest above the other. (b) Top-1 success in each group, with 95\% confidence intervals from a paired PDB-entry bootstrap ($^{***}$two-sided $p \leq 0.001$).}
\label{fig:si_core_vs_min_rmsd}
\end{figure}

\newpage

\subsection*{S7. Reliability analyses retaining invalid poses}
These sensitivity analyses retain all original poses for clustering, structural summaries, and confidence-based selection. Success still requires all three checks: RMSD below $2$~\AA, lDDT-PLI above $0.8$, and PoseBusters-validity. All eight displayed measures use the same 12,990 ensembles within their respective models. This is a separately matched cohort; differences from the valid-only analysis reflect both candidate filtering and eligibility, so differences in AUC are not a paired estimate of the filtering effect.

\begin{figure}[H]
\centering
\includegraphics[width=\textwidth]{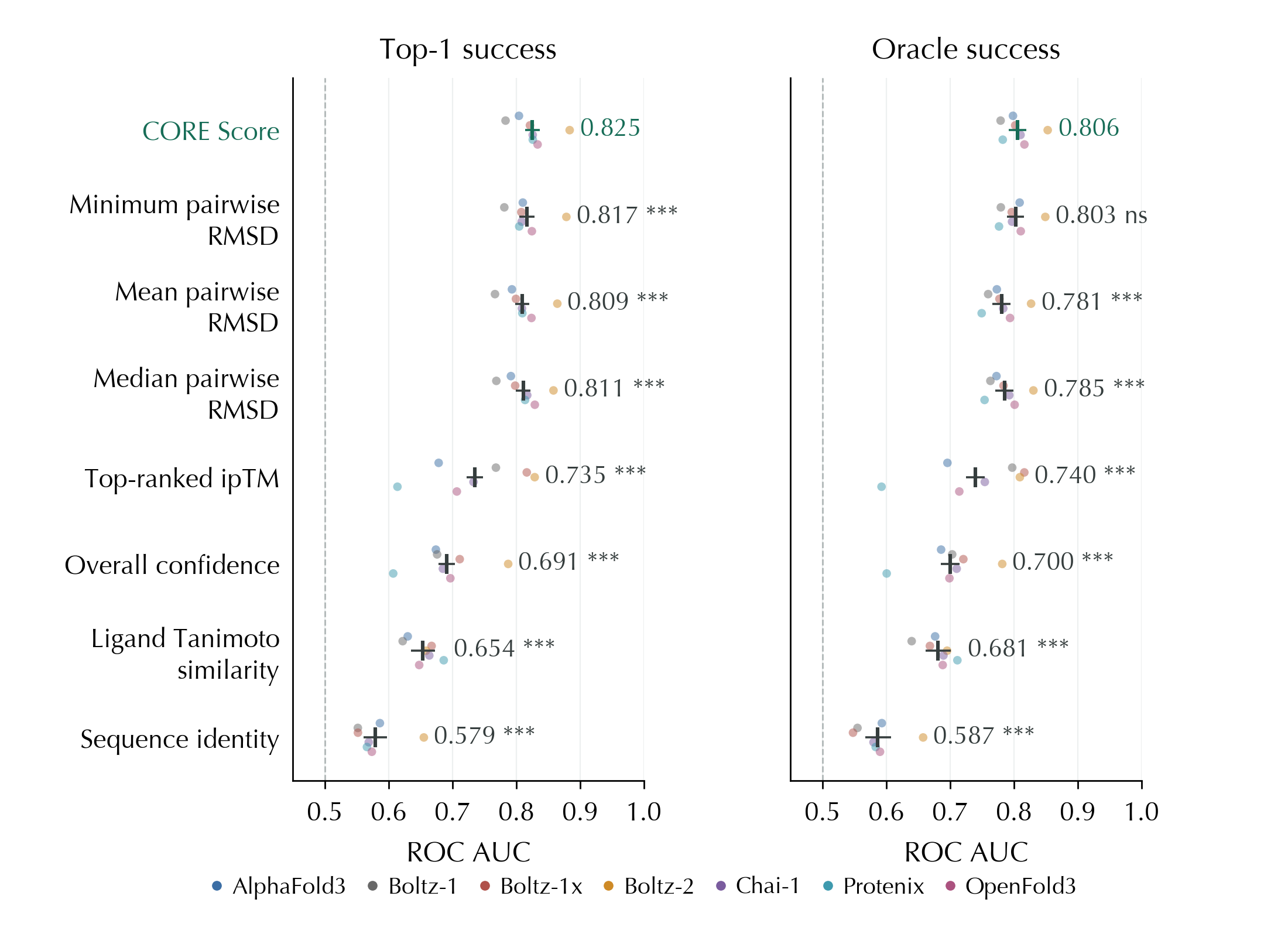}
\caption{\textbf{ROC AUC with invalid poses retained.} Colours identify models. Black marks show equal-model means with 95\% confidence intervals, and asterisks mark differences from CORE Score in equal-model mean ROC AUC ($^{***}$two-sided $p \leq 0.001$), each from the same paired PDB-entry bootstrap. Overall confidence evaluates its own selected pose, while the other measures evaluate the highest-ipTM pose.}
\label{fig:si_allposes_auc}
\end{figure}

\begin{figure}[H]
\centering
\includegraphics[width=\textwidth]{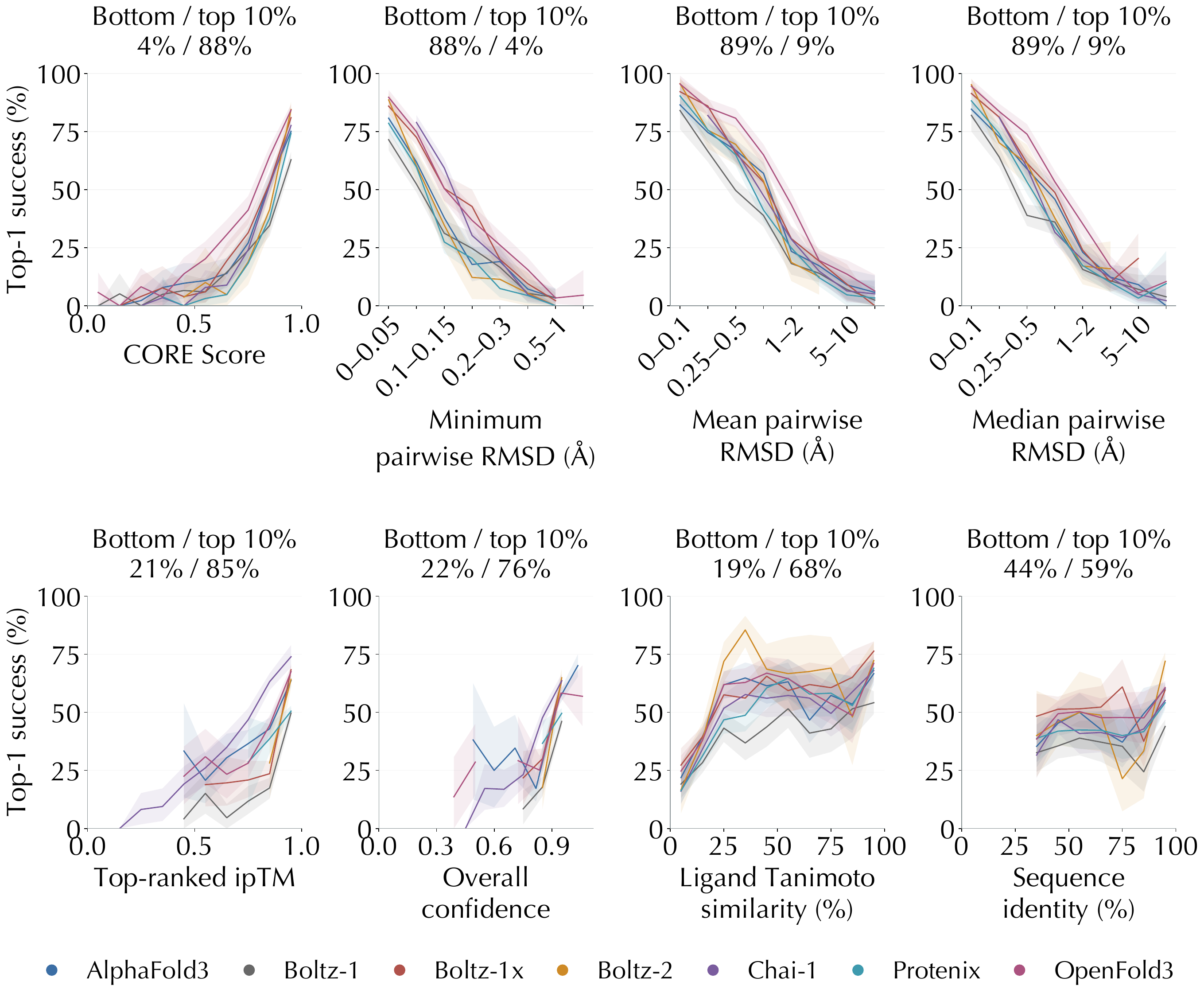}
\caption{\textbf{Score-success relationships with invalid poses retained.} These curves use the same cohort as Figure~\ref{fig:si_allposes_auc}. Bands show pointwise 95\% PDB-entry bootstrap intervals. Annotations report mean success in the lowest and highest scoring tenths across models. Bins with fewer than 20 ensembles are omitted.}
\label{fig:si_allposes_curves}
\end{figure}

\newpage

\subsection*{S8. Stricter success thresholds}
We repeated the analysis with success defined by ligand RMSD below $R = 0.5$, $1$, and $1.5$~\AA, keeping lDDT-PLI above $0.8$ and PoseBusters validity, with $T$ set equal to $R$. As the threshold became stricter, CORE Score discrimination of top-1 success increased from 0.866 at $2$~\AA\ to 0.885 at $0.5$~\AA, while ipTM decreased from 0.757 to 0.728. The difference between CORE Score and ipTM therefore widened from 0.109 to 0.157. CORE Score remained the strongest measure at every threshold and for both outcomes. 

\begin{table}[H]
\centering
\small
\caption{Mean ROC AUC for top-1 success across success thresholds $R$.}
\label{tab:si_threshold}
\begin{tabular}{lrrrr}
\hline
Measure & $0.5$~\AA & $1$~\AA & $1.5$~\AA & $2$~\AA \\
\hline
CORE Score ($T = R$) & \textbf{0.885} & \textbf{0.874} & \textbf{0.867} & \textbf{0.866} \\
Median pairwise RMSD & 0.872 & 0.860 & 0.851 & 0.849 \\
Minimum pairwise RMSD & 0.873 & 0.859 & 0.851 & 0.850 \\
Top-ranked ipTM & 0.728 & 0.744 & 0.753 & 0.757 \\
Runs N' Poses similarity & 0.773 & 0.765 & 0.767 & 0.771 \\
Ligand Tanimoto similarity & 0.695 & 0.686 & 0.686 & 0.689 \\
Sequence identity & 0.538 & 0.562 & 0.575 & 0.580 \\
\hline
\end{tabular}
\end{table}

Because $T$ was set equal to $R$ above, the contributions of the two thresholds cannot be separated. We therefore repeated the analysis over a grid of both, varying $T$ from $0.5$ to $3$~\AA\ at success thresholds $R$ from $0.5$ to $3$~\AA\ (Table~\ref{tab:si_threshold_grid}).

\begin{table}[H]
\centering
\caption{Mean ROC AUC for top-1 success with the CORE Score threshold $T$ varied independently of the success threshold $R$. Underlined values are those with $T = R$, as reported in Table~\ref{tab:si_threshold}.}
\label{tab:si_threshold_grid}
\begin{tabular}{lcccccc}
\hline
 & \multicolumn{6}{c}{CORE Score threshold $T$ (\AA)} \\
Success threshold $R$ & $0.5$ & $1$ & $1.5$ & $2$ & $2.5$ & $3$ \\
\hline
$0.5$~\AA & \underline{0.885} & 0.882 & 0.880 & 0.878 & 0.878 & 0.877 \\
$1$~\AA & 0.875 & \underline{0.874} & 0.873 & 0.872 & 0.872 & 0.871 \\
$1.5$~\AA & 0.866 & 0.867 & \underline{0.867} & 0.867 & 0.867 & 0.866 \\
$2$~\AA & 0.864 & 0.866 & 0.866 & \underline{0.866} & 0.867 & 0.866 \\
$2.5$~\AA & 0.863 & 0.864 & 0.865 & 0.865 & \underline{0.865} & 0.865 \\
$3$~\AA & 0.862 & 0.863 & 0.864 & 0.865 & 0.865 & \underline{0.865} \\
\hline
\end{tabular}
\end{table}

\newpage

\subsection*{S9. Structural diversity and modes before PB filtering}
These analyses use all original poses and RMSD below $2$~\AA\ and lDDT-PLI above $0.8$ as the success criteria.

\begin{figure}[H]
\centering
\includegraphics[width=\textwidth]{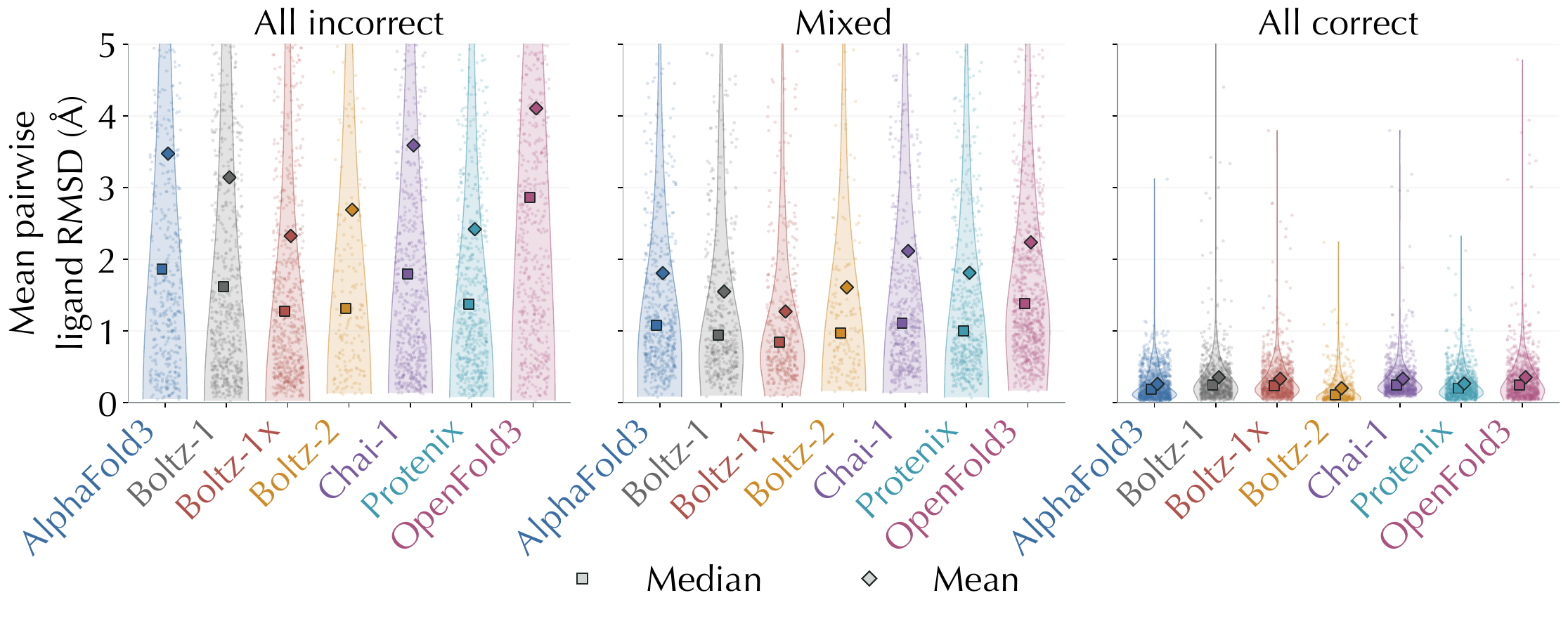}
\caption{\textbf{Diversity distributions by structural outcome across all seven models.} All original poses are included. Dots show model-specific ensembles, violins show their distributions, squares show medians, and diamonds show means. The displayed range is 0-5~\AA; summary statistics use every observation, including those outside this view. The complete range is shown in Figure~\ref{fig:si_diversity_full}.}
\label{fig:si_diversity}
\end{figure}

\begin{figure}[H]
\centering
\includegraphics[width=\textwidth]{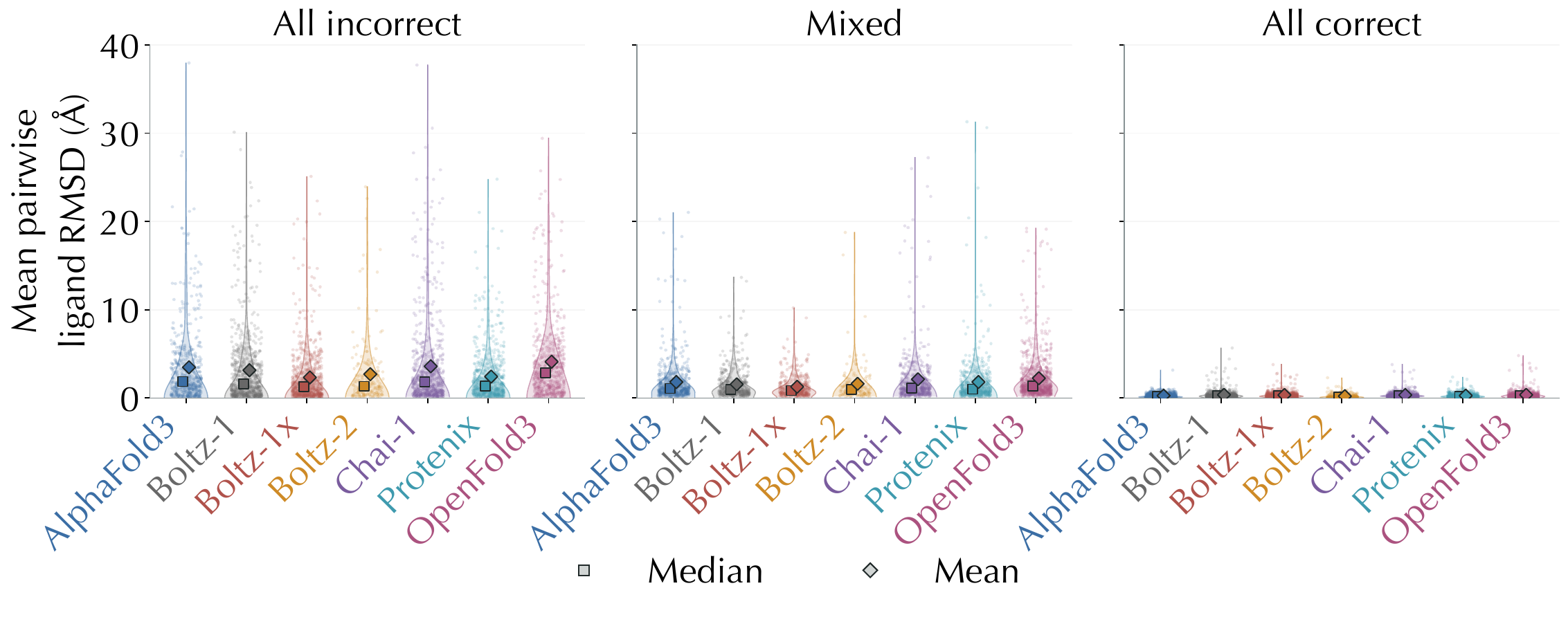}
\caption{\textbf{Full-range diversity distributions.} The observations, colours and summary markers are identical to Figure~\ref{fig:si_diversity}; all observed RMSD values are visible.}
\label{fig:si_diversity_full}
\end{figure}

\begin{figure}[H]
\centering
\includegraphics[width=\textwidth]{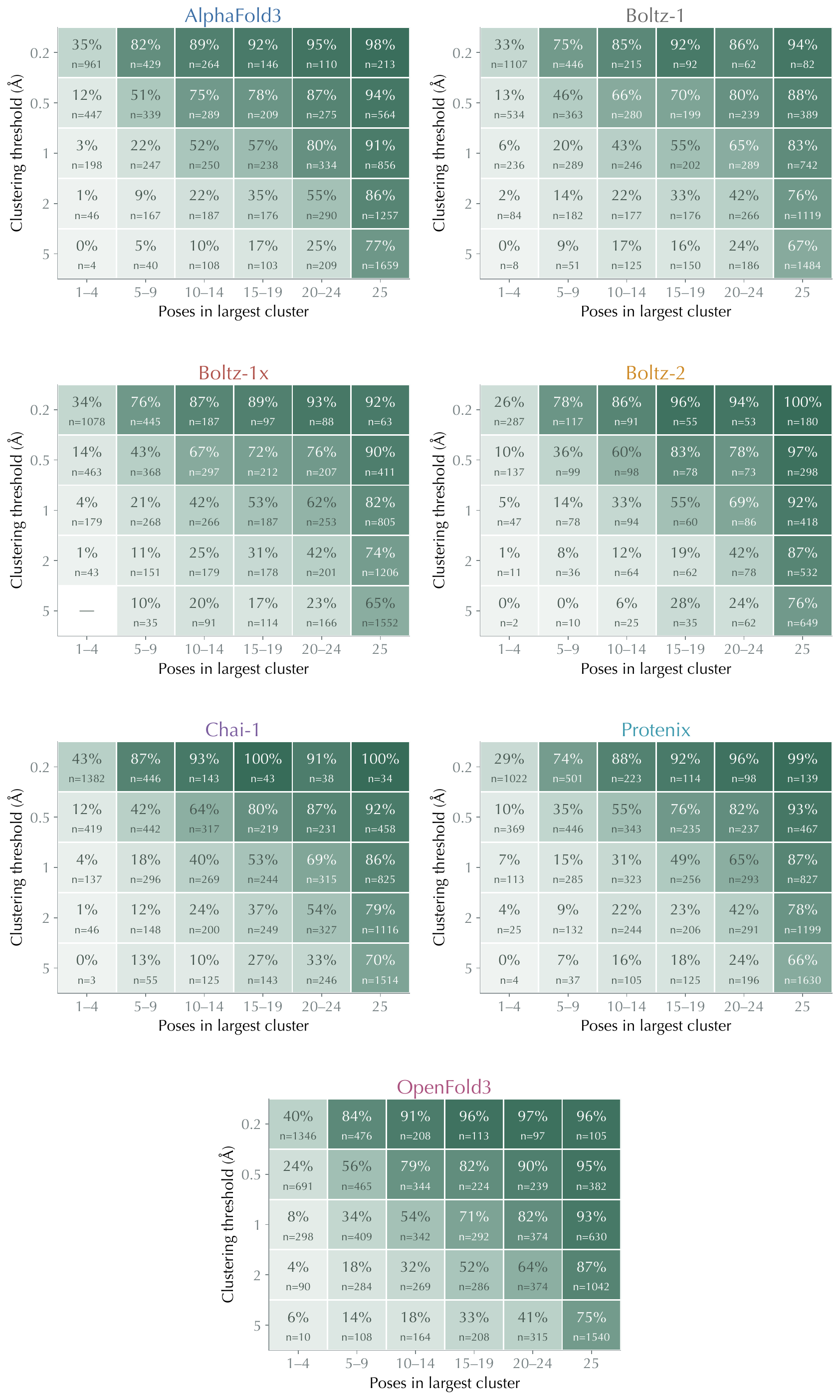}
\caption{\textbf{Largest-mode correctness increases with size and tighter agreement.} Model-specific percentages of structurally correct poses in the largest complete-linkage mode, grouped by size and ligand RMSD cutoff. Tied largest modes are averaged within each ensemble. Each cell gives the number of ensembles, $n$; a dash marks no observations.}
\label{fig:si_mode_size}
\end{figure}

\newpage

\subsection*{S10. Oracle success}
Oracle success requires at least one PB-valid pose in the ensemble with RMSD below $2$~\AA\ and lDDT-PLI above $0.8$, and so does not depend on which pose is selected. CORE Score achieved an average ROC AUC of 0.838 for oracle success, greater than all other metrics compared here.

\begin{figure}[H]
\centering
\includegraphics[width=\textwidth]{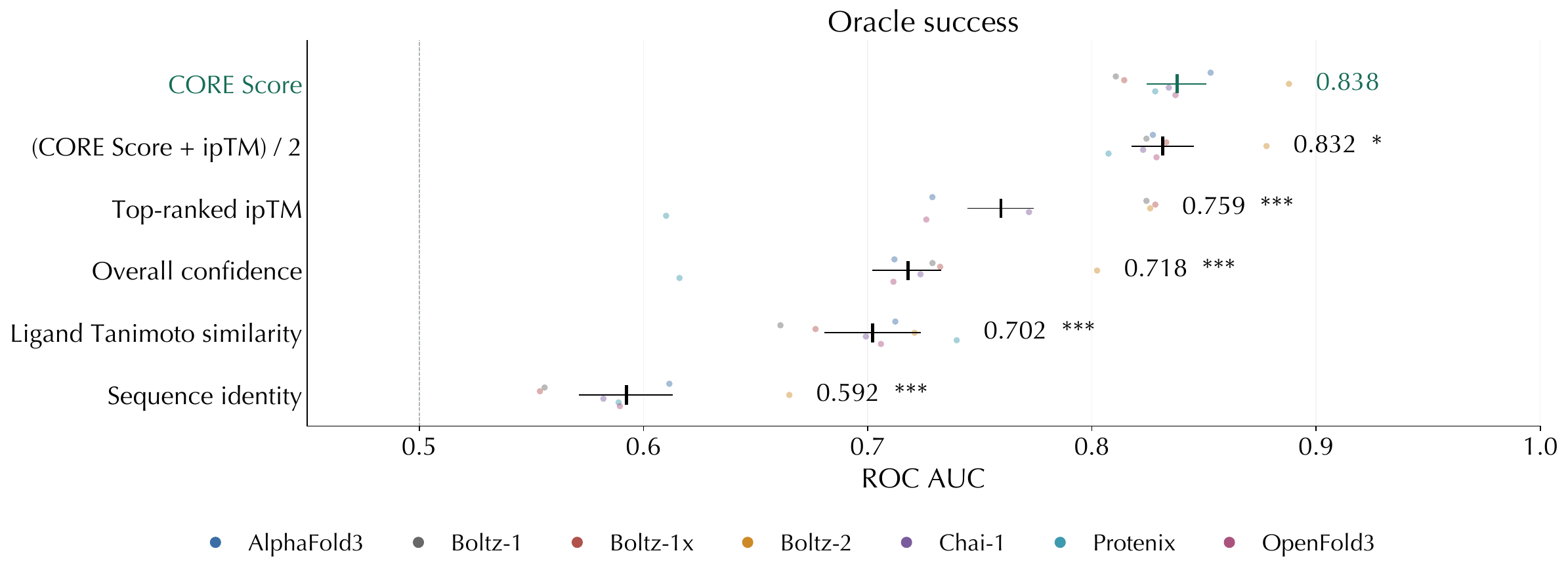}
\caption{\textbf{Discrimination of oracle success.} ROC AUCs for whether any PB-valid pose in the ensemble is correct. Coloured points identify models. Black marks show equal-model means with 95\% confidence intervals, and asterisks mark differences from CORE Score in equal-model mean ROC AUC ($^{***}$two-sided $p \leq 0.001$), each from the same paired PDB-entry bootstrap.}
\label{fig:si_oracle_auc}
\end{figure}

\newpage

\subsection*{S11. Retaining predictions by score}

A score is used prospectively by keeping the highest-scoring predictions and discarding the rest. Ensembles were therefore ranked within each model and the outcome recalculated among the retained and the discarded fractions. Retaining the highest-scoring half of predictions, top-1 success rose from 61.3\% to 87.3\% for CORE Score, compared with 79.2\% for top-ranked ipTM. At a retained fraction of 10\%, the corresponding rates were 96.1\% and 92.0\%. The separation was larger among discarded predictions. Among the lowest-scoring tenth, 94.2\% of top-1 predictions were incorrect by CORE Score, compared with 72.7\% by ipTM, against a background failure rate of 38.7\%.

\begin{figure}[h]
\centering
\includegraphics[width=\textwidth]{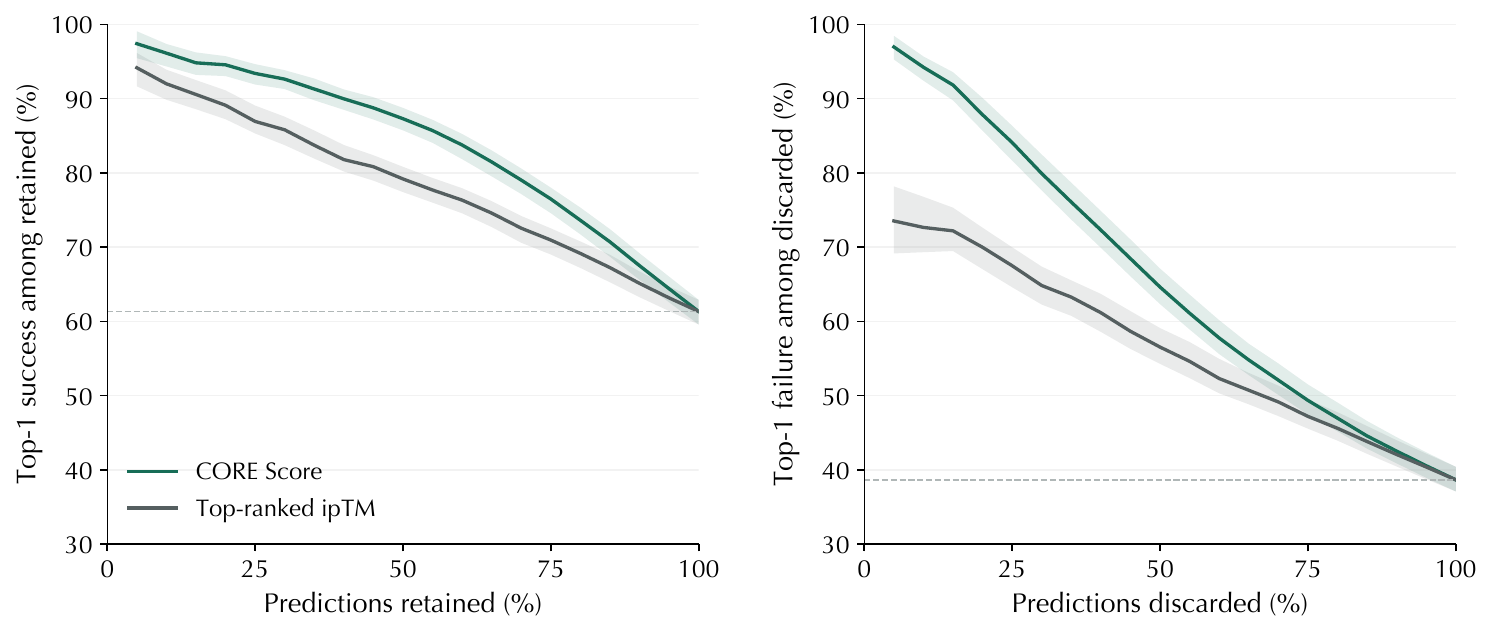}
\caption{\textbf{Outcome among retained and discarded predictions.} Left: top-1 success among the highest-scoring fraction retained. Right: top-1 failure among the lowest-scoring fraction discarded. Predictions are ranked within each model. Lines show equal-model means and bands show 95\% PDB-entry bootstrap intervals. Dashed lines mark the rates with all predictions included.}
\label{fig:si_retention}
\end{figure}

\newpage

\subsection*{S12. Structural similarity within individual models}

\begin{figure}[H]
\centering
\includegraphics[width=\textwidth]{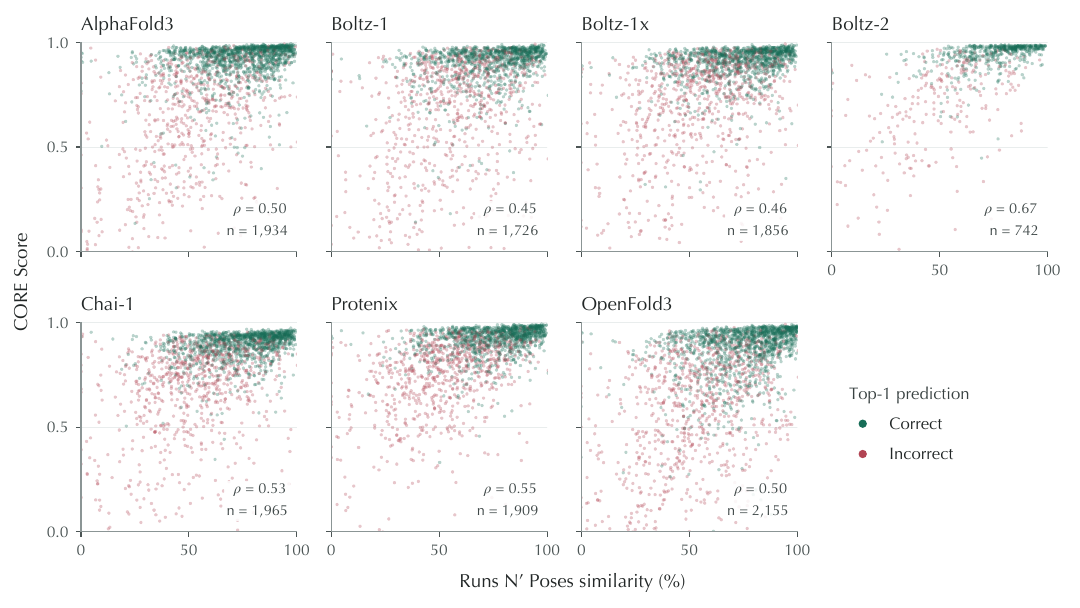}
\caption{\textbf{CORE Score and structural similarity within each model.} These panels separate the main pooled scatter into its seven model-specific cohorts, using the same 12,287 ensembles. Colours show correctness of the highest-ipTM valid pose. Annotations give Spearman correlations and model-specific sample sizes.}
\label{fig:si_s3_correlation_models}
\end{figure}

\begin{figure}[H]
\centering
\includegraphics[width=\textwidth]{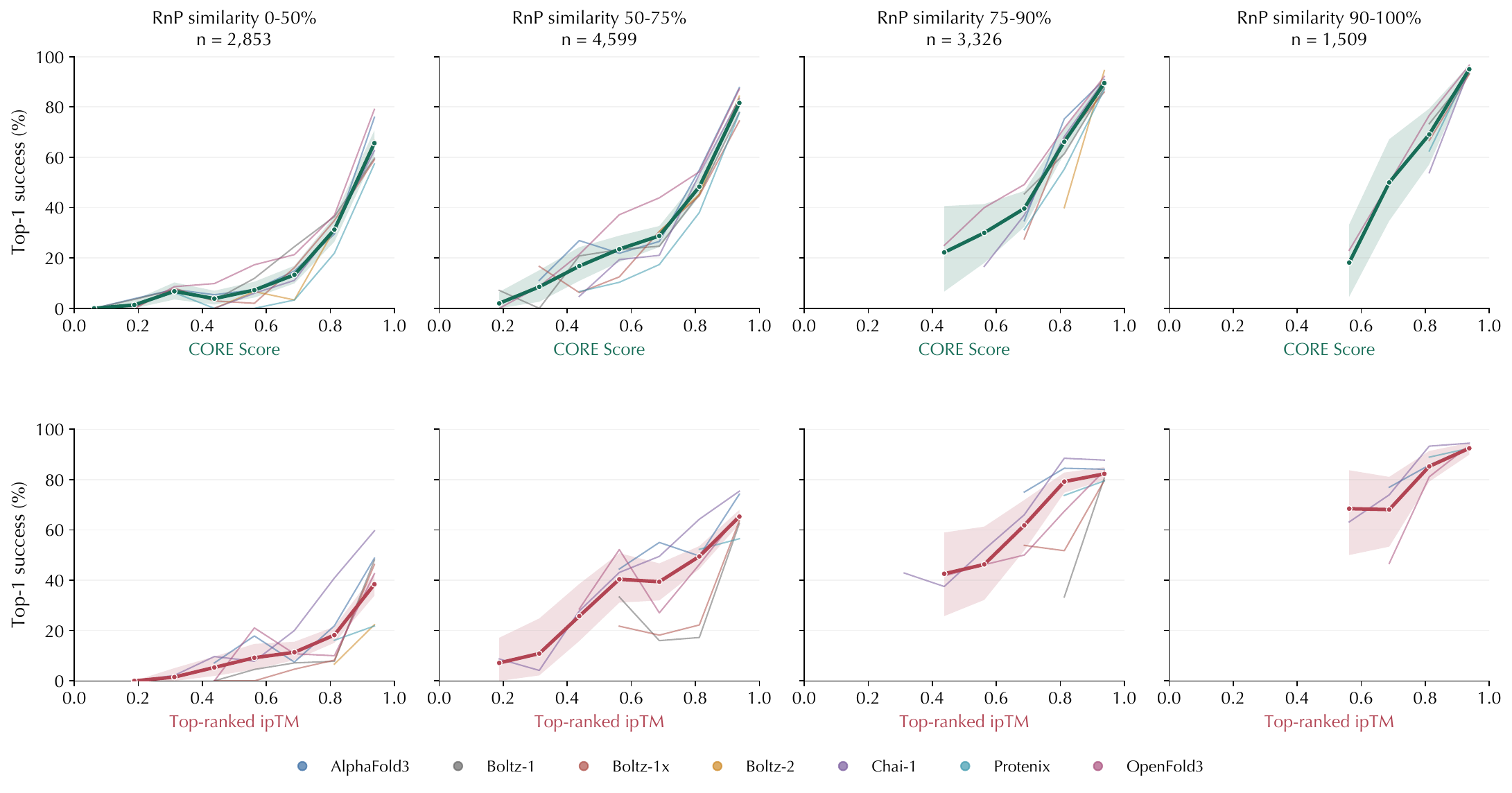}
\caption{\textbf{Score and top-1 success within Runs N' Poses similarity strata.} Thick lines show the mean across all ensembles in each stratum with 95\% PDB-entry bootstrap bands. Thin lines show individual models.}
\label{fig:si_curves_by_similarity}
\end{figure}

Discrimination was also examined within similarity strata. Ensembles were grouped by Runs N' Poses similarity and the relationship between score and top-1 success was recalculated within each group (Figure~\ref{fig:si_curves_by_similarity}). CORE Score remained associated with top-1 success in every stratum, with mean ROC AUCs of 0.853, 0.810, 0.812, and 0.860 from the least to the most similar group, compared with 0.801, 0.728, 0.703, and 0.761 for ipTM.

\newpage

\subsection*{S13. Example targets across all models}

\begin{figure}[H]
\centering
\includegraphics[width=\textwidth]{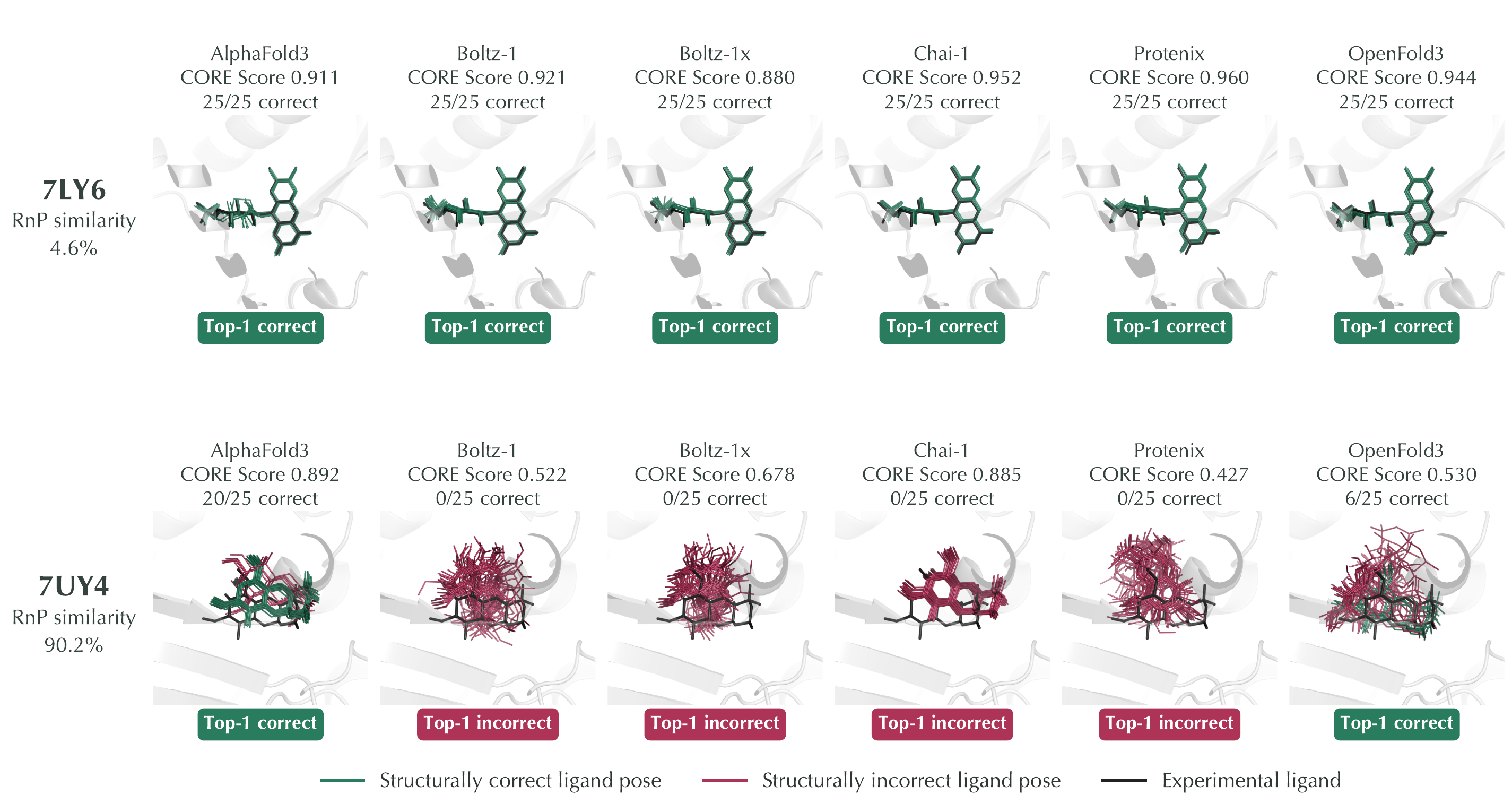}
\caption{Predicted ligand poses for 7LY6 and 7UY4 from each model, overlaid on the experimental ligand (black).}
\label{fig:si_examples_all_models}
\end{figure}

Figure~\ref{fig:si_examples_all_models} shows the predictions for 7LY6 and 7UY4 from every model. For 7LY6, all models produced 25 of 25 correct poses and a correct top-1 pose. For 7UY4, AlphaFold3 and OpenFold3 produced correct poses and a correct top-1 pose, while the other four models produced none. Chai-1 produced a concentrated ensemble just outside the success thresholds (median RMSD 2.23~\AA{}), giving a high CORE Score of 0.885 despite no correct poses.

\end{document}